\documentclass[10pt,journal,compsoc]{IEEEtran}
\usepackage[modulo,switch]{lineno}

\ifCLASSOPTIONcompsoc
  \usepackage[nocompress]{cite}
\else
  \usepackage{cite}
\fi
\ifCLASSINFOpdf
\else
\fi

\usepackage{times}
\usepackage{epsfig}
\usepackage{graphicx}
\usepackage{amsmath}
\usepackage{amssymb}
\usepackage{multirow}
\usepackage{subfig}
\usepackage{amsthm}
\usepackage{bbm}
\usepackage[ruled,vlined]{algorithm2e}
\usepackage{makecell}
\usepackage{rotating}
\usepackage{booktabs}
\usepackage{float}
\usepackage{color}
\usepackage[modulo,switch]{lineno}

\makeatletter
\newcommand{\removelatexerror}{\let\@latex@error\@gobble}
\makeatother

\newcommand{\scal}[1]{\mathit{#1}}
\newcommand{\vect}[1]{\mathbf{#1}}
\newcommand{\matr}[1]{\mathbf{#1}}

\newcommand{\set}[1]{\mathcal{#1}}

\newcommand{\changed}[1]{\textcolor{black}{#1}}

\newcommand{\ra}[1]{\renewcommand{\arraystretch}{#1}}

\begin{document}

%
\title{{3D Rigid Motion Segmentation with Mixed and Unknown Number of Models}}
%
%
%

\author{Xun~Xu,~\IEEEmembership{Member,~IEEE},
        Loong-Fah~Cheong,
        and~Zhuwen~Li,
\IEEEcompsocitemizethanks{\IEEEcompsocthanksitem 
X. Xu and  L.-F. Cheong are both with the Department of Electrical and Computer Engineering, National University of Singapore, Singapore\protect\\
E-mail: alex.xun.xu@gmail.com \& eleclf@nus.edu.sg\protect\\
Z. Li is with the PonyAI, US\protect\\
E-mail: lzhuwen@gmail.com
}
\thanks{}}

%
%

\markboth{Journal of \LaTeX\ Class Files,~Vol.~14, No.~8, August~2015}%
{Shell \MakeLowercase{\textit{et al.}}: Bare Demo of IEEEtran.cls for Computer Society Journals}
%



\IEEEtitleabstractindextext{%
\begin{abstract}
Many real-world {video sequences} cannot be conveniently categorized as general or degenerate; in such cases, imposing a false dichotomy in using the fundamental matrix or homography model for motion segmentation {on video sequences} would lead to difficulty. Even when we are confronted with a general scene-motion, the fundamental matrix approach as a model for motion segmentation still suffers from several defects, which we discuss in this paper. The full potential of the fundamental matrix approach could only be realized if we judiciously harness information from the simpler homography model. From these considerations, we propose a {multi-model spectral clustering} framework that synergistically combines multiple models {(homography and fundamental matrix)} together. We show that the performance can be substantially improved in this way. For general motion segmentation tasks, the number of {independently moving objects} is often unknown a priori and needs to be estimated from the observations. This is referred to as model selection and it is essentially still an open research problem. In this work, we propose a set of model selection criteria balancing data fidelity and model complexity.
We perform extensive testing on existing motion segmentation datasets with both segmentation and model selection tasks, achieving state-of-the-art performance on all of them; we also put forth a more realistic and challenging dataset adapted from the KITTI benchmark, containing real-world effects such as strong perspectives and strong forward translations not seen in the traditional datasets.
\end{abstract}

}

\maketitle

\IEEEdisplaynontitleabstractindextext

%
\IEEEpeerreviewmaketitle

\begingroup

\IEEEraisesectionheading{\section{Introduction}\label{sec:introduction}}

%
%
%
%

\IEEEPARstart{T}{he} {task of 3D motion segmentation aims to separate tracked feature points (in our case a sparse set of trajectories in a video sequence) according to the respective rigid 3D motion.} Various geometric models have been used in the 3D motion segmentation problem to model the different types of cameras, scenes, and motion. In this problem as commonly set forth, the underlying models are generally regarded as applicable under different scenarios and these scenarios do not overlap. For instance, when the underlying motion contains translation and the scene structure is non-planar, the fundamental matrix is used to model the epipolar geometry \cite{Jung2014a,Li2013}. \changed{When the scene-motion is degenerate, i.e. close-to-planar structure and/or vanishing camera translation, the homography is preferred \cite{Dragon2012,Lai2017}.} 
However, the real world scene-motions are in fact not so conveniently divided. \changed{They are more typified by near-degenerate scenarios such as a scene that is almost but not quite planar, or a motion that is rotation-dominant but with a non-vanishing translation.} In such cases, imposing a false dichotomy in deciding an appropriate model would pose difficulty for subsequent subspace separation. For instance, it is well-known \cite{Goshen2008,sugaya2004geometric,Torr1998} in the case of a scene with dominant-plane, it is easy to find inliers belonging to the degenerate configuration (the plane), but the precision of the resulting fundamental matrix is likely to be very low. Most of the inliers outside the degenerate configuration will be lost, and often the erroneous fundamental matrix will pick up outliers (e.g. from other motion groups). Since this is not a purely planar scene, using homography in a naive manner might fail to group all the inliers together too, resulting in over-segmentation of the subspaces.

It is also not hard to establish---from a glance of the motion segmentation literature---that of the various models, the fundamental matrix model is generally eschewed, due to the lack of perspective effects in the Hopkins155 benchmark \cite{Tron2007}. However, it is never clearly articulated if the numerical difficulties arising from degeneracies in such approach present insuperable obstacles. And no one has put his/her finger on the exact manner how the resulting affinity matrix is ill-suited for subspace clustering: is it solely due to the degeneracies or are there other factors? Considering that in many real-world applications say, autonomous driving, perspective effects are not uncommon, it surely follows that we should come to a better understanding of the suitability of fundamental matrix (or for that matter, the homography model) as a geometric model for motion segmentation. This, we contend, is far from being the case. {For instance, does it follow that if we use the fundamental matrix for wide field-of-view scenes {(such that strong perspective effects exist \cite{hartley2003multiple})}, like those found in the KITTI benchmark \cite{Geiger2013IJRR}, we will get better performance than those using homography? We have in fact as yet no reason to believe that this will be the case, judging by the way how the best performing algorithm is based on homography model \cite{Lai2017}, outperforming those based on fundamental matrix \cite{Li2013}. \changed{ Empirically, we observed that this superiority still persists for the MTPV62 dataset and individual Hopkin sequences that have larger perspectives (though admittedly still moderate in the latter).} Indeed, from the results we obtained on the KITTI sequences that we adapted for testing motion segmentation in real-world scenarios, the superiority of the homography-based methods is again observed.} Thus, one might naturally ask what factors other than degeneracies are hurting the fundamental matrix approach? And why is the homography matrix approach holding its own in wide perspective scenes, when it possesses none of the geometrical exactness of the fundamental matrix?

In the remainder of this section, we will briefly investigate the suitability of homography and fundamental matrices ($\mathbf{H}$ and $\mathbf{F}$ respectively) as a geometric model for motion segmentation. We shall henceforth denote the affinity matrices generated by $\mathbf{H}$ and $\mathbf{F}$ as $\mathbf{K_H}$ and $\mathbf{K_F}$ respectively.

\subsection{Success Roadmap of $\mathbf{H}$}

{We} have already alluded to the fact that the affinity matrix $\mathbf{K_H}$ may not exhibit high intra-cluster cohesion (due to lack of strong affinity between different planes of the same rigid motion) and is inadequate for 3D motion segmentation. 
 {In the Hopkins155 dataset, this is not an overriding concern since most of the sequences have a small field-of-view and are dominated by pure camera rotation. 
 These are seemingly evidenced by the good empirical results obtained by a wide variety of approaches based on affine subspace or homography matrix.} The recent homography-based method \cite{Lai2017} boasts state-of-the-art performance with a mean error of $0.83\%$. The low error attained is noteworthy given that there are actually some Hopkins sequences with non-negligible perspective effects and significant camera translation (induced by self-rotating objects); we feel that this phenomenon warrants a better explanation than the reasoning offered so far.

{As we observe from the real hypotheses shown in Fig.~\ref{fig:planar_slice} (a-b),} the success can be attributed to the many planar slices induced by the homography hypothesizing process; these are not necessarily actual physical planes in the scenes (see the slices in Fig.~\ref{fig:planar_slice} (a-b)) but as long as these virtual planes belong to the same rigid motion, it is evident that they can be fitted with a homography. Such slicings of the scene create strong connections between points across multiple real planar surfaces and result in a much less over-segmented affinity matrix $\mathbf{K_H}$. If the scene contains only compact objects or piecewise smooth structures, then such connectivity created is sufficient to bind the various surfaces of a rigid motion together. However, in the real world sequences, when the above conditions are not satisfied, we suspect that this may not be adequate. Fig.~\ref{fig:planar_slice}(c) illustrates a background comprising an elongated object (a traffic light) and the marking on the road. It is clear that in this case, while one can form virtual planar slices as before, the resulting connectivity is much lower (most if not all of the slices cannot connect large segments of both these elements simultaneously, unlike those in Fig.~\ref{fig:planar_slice} (a-b)). 

\begin{figure}[!ht]
\begin{center}
\includegraphics[width=0.95\linewidth]{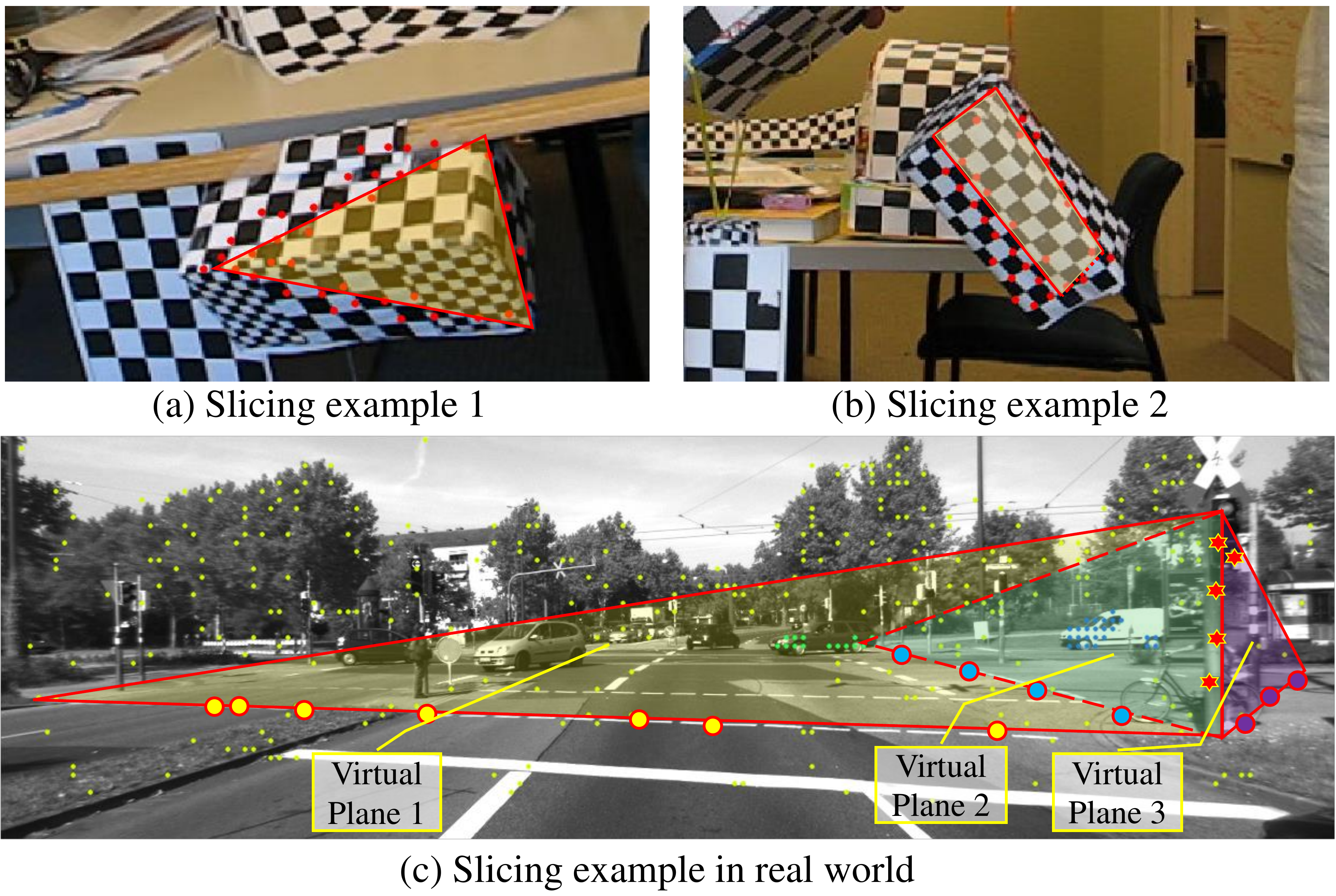}
\caption{Illustration of slicing effect of homography. (a-b) Red dots indicate inlier points of a hypothesis. All points lie on a virtual plane (a slice of the cube) highlighted in yellow. (c) Virtual planes are highlighted as triangles with points in the same color as inliers. } 
\label{fig:planar_slice}
\end{center}
\end{figure}

\subsection{Problems with $\mathbf{F}$}

Besides the degeneracy issues, another root problem
with the fundamental matrix approach for the motion segmentation
problem lies precisely in the fact that {it is a complicated model that
becomes susceptible to capturing fictitious scene-motion configurations. For example, when the 8 background points are not well chosen (e.g. occupying a small spatial extent, or residing on a surface close to a plane), the estimated \textbf{F} can be quite far from the true \textbf{F}. This is the case for the scenes depicted in Fig.~\ref{fig:F_encompassing}: the fitted \textbf{F} erroneously incorporates points on the moving foreground but misses some stationary background points.}

\begin{figure}[!hb]
\includegraphics[width=1.01\linewidth]{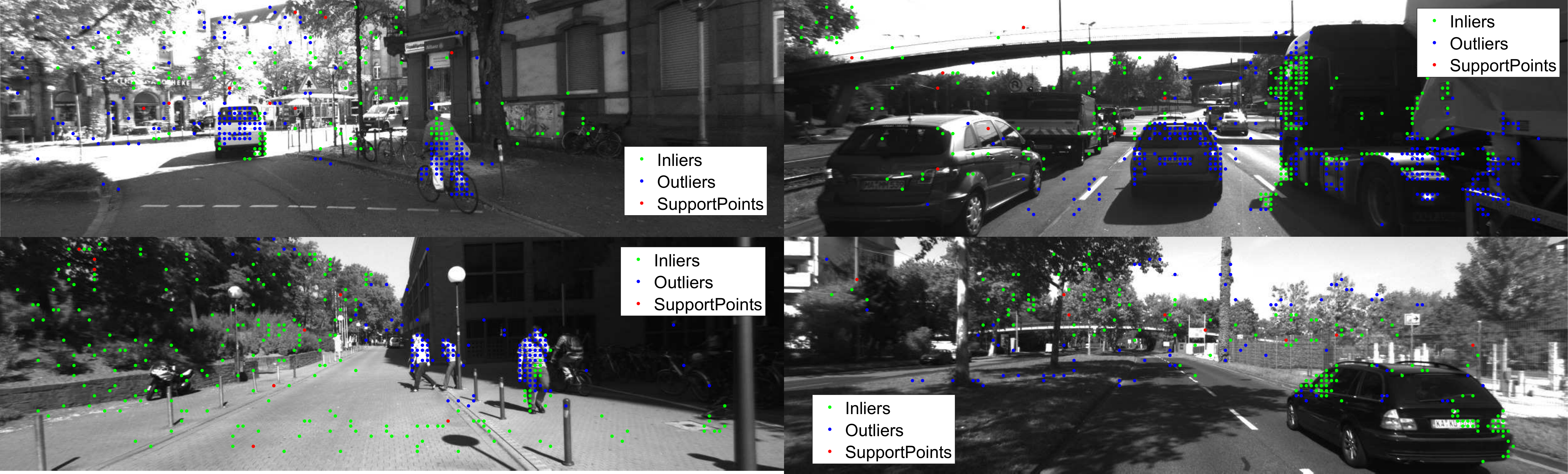}
\caption{{Hypothesized \textbf{F} from background erroneously captures foreground as inliers.}}\label{fig:F_encompassing}
\end{figure}

Using $\matr{F}$ alone renders it likely to capture any correlation between different rigid motion groups. Therefore, compared to a simpler model such as homography, it is more likely to cause overlapping between the subspaces of different rigid motion groups. 
However, is it not possible that $\matr{F}$ also offers the greatest scope for forming the best {correct model}, given that it starts with a geometrically correct model and must have thus captured much of what is correct? It perhaps requires some nudge in the correct direction for us to reclaim the performance that ought be had for $\mathbf{K_F}$. From this standpoint, even when we are confronted with a general scene-motion with no degeneracies, there is still an important 
reason for keeping the homography model---to midwife the unborn model of $\mathbf{K_F}$.

\subsection{Model Selection}
Choosing the type of geometric model to describe the motion is one of the open challenges in motion segmentation. Another open challenge in motion segmentation is to automatically determine the number of moving objects, which is also referred to as model selection in the literature \cite{Tibshirani2001,Chin2010,Li2013,Li2017,Lai2017}. It is well known that with perfect block diagonal affinity matrix, the number of cluster corresponds to exactly the multiplicities of the zero eigenvalue of the Laplacian matrix \cite{von2007tutorial}. However, such noise-free affinity matrix rarely exists in real-world problems. The modified rule of gap heuristic \cite{von2007tutorial} is purported to deal with noisy affinity matrix but has seldom worked well in practice. Those approaches that have performed well for model selection tend to be those that explicitly acknowledge the presence of noise and either seek its removal like LRR \cite{liu2013robust} or repair the affinity matrix like SCAMS \cite{Li2014}. We too believe that it is important to avail ourselves of this flexibility to deviate from the observed affinity matrix. Putting this criterion together with the classical criteria of good clusters, we arrive at the following three desiderata for model selection and membership assignment: 1) the affinity reconstructed from the clustering partition should be close to the original affinity matrix; 2) the intra-cluster similarity should be maximized; and 3) the inter-cluster similarity should be minimized. As is well known \cite{von2007tutorial}, the latter two criteria are encapsulated in the objective function for normalized cut. The first criterion can be regarded as the difference between original and reconstructed affinities, namely the reconstruction error; thus we name our model selection approach as normalized cut with reconstruction error (NCRE). As simple as this extension seems, the research community has surprisingly yet to explore, as far as we know, all of these criteria in an integral manner in the form of a global objective function. Later in this paper, we will
show how our normalized reconstruction error term can counterbalance the tendency of the normalized cut term to under-segment clusters.

\subsection{Proposed Solution}
We 
point out that many real-world sequences cannot be classified into neat categories such as general or degenerate scene-motions and thus cannot be adequately addressed by any single model such as $\mathbf{H}$ or $\mathbf{F}$. We have also discussed the defects of the fundamental matrix approach and conjectured that 
its full potential could perhaps be realized if we judiciously harness information from a simpler model such as $\mathbf{H}$. From these considerations, we propose a multi-model
 spectral clustering framework that synergistically combines these multiple models together. As there is no definite consensus on how best to combine {several models} together for spectral clustering, we evaluate a few extant fusion schemes. Since these generic schemes ignore the hierarchical relationship of the affinity matrix in each model, we also put forth a custom-made fusion scheme that preserves the true hierarchical structure of the affinity matrices.
By doing so, we make sure that our findings are not an artifact of a particular fusion scheme. As we will show later, the performance of the fundamental matrix approach can be substantially raised using the improved $\mathbf{K_F}$.
We hasten to add that one should not over-claim the potential gains of this fundamental matrix approach. When the scene contains substantial amount of degeneracies
 it is always better to rely on the combined model for the best performance. 
To enable automatic selection of the number of moving objects, we propose a novel set of model selection criteria.

To summarize, the contributions of our paper are as follows. First, we contribute to an understanding of the strengths and drawbacks of homography and fundamental matrices as a geometric model for motion segmentation. We then propose using affinity matrix fusion as a means of dealing with real-world effects that are often difficult to model with a pure homography or fundamental matrix. We also propose a novel model selection criteria balancing a good fit to the data and model complexity. 
Finally, we perform extensive testing on existing motion segmentation datasets, achieving state-of-the-art performance on all of them; we also put forth a more realistic and challenging dataset adapted from the KITTI benchmark, containing real-world effects such as strong perspectives and strong forward translations not seen in the traditional datasets.

\section{Related Work}

In this section, we review related literature {from three perspectives}. We first review the recent advances in 3D motion segmentation. Then we briefly review representative works in multi-view unsupervised clustering. Finally, we discuss the recent works in model selection with focus on motion segmentation task.

\noindent\textbf{Motion Segmentation} 
Research into 3D motion segmentation can be divided into two major groups: those based on a hypothesis-and-test paradigm and those that are more analytic rather than hypothesis-driven. Into the latter camp falls a wide variety of approaches, including factorization \cite{boult1991factorization,costeira1998multibody, gear1998multibody,gruber2004multibody,tomasi1992shape}, algebraic method \cite{Rao2010,vidal2004motion,vidal2005generalized,Vidal2008}, affinity matrix \cite{lauer2009spectral,yan2006general}, including those constructed from sparse representation \cite{Elhamifar2013,liu2013robust}. They typically assume that the input is made up of a union 
of motion groups of specific types, with only a few works \cite{goh2007segmenting,Rao2010} that can handle mixed types of motion groups. These analytic approaches are rightly praised for their elegance but become awkward in dealing with real world signals that are often drawn from mixed multiple manifolds. In contrast, works in the former category, being hypothesis-driven, are naturally more suited to handling mixed models. This is exemplified in the earlier works such as \cite{sugaya2004geometric,Torr1998} which explicitly decide on whether $\mathbf{F}$ or $\mathbf{H}$ is better suited as a motion model in the face of possibly degenerate scene-motion configuration, but these works are applied to cases where the background is by far the most dominant group in the scene. 
Subsequent hypothesis-and-test methods \cite{Chin2009,chin2010accelerated,lazic2009floss} dealing with the realistic Hopkins155 \cite{Tron2007} sequences almost as a rule ignore the more complex fundamental matrix 
altogether.
Thus, these later works do not concern themselves with the problem of dealing with mixed types of models. Our approach differs from the above works in that not only do we allow for mixed types of models, we also do not impose a dichotomous decision on what is an appropriate model.

{Finally, we briefly mention works based on alternative definitions of motion segmentation \cite{ochs2014segmentation,keuper2015motion,keuper2017higher, bideau2016s,bideau2018best,zamalieva2014background}. These works segment motion groups entirely in the 2D domain, and as such, cannot deal with the motion discontinuities arising from background surfaces residing at different depths. Some works such as \cite{bideau2016s,zamalieva2014background} do not attempt to separate the differently moving foreground objects, being interested only in a binary foreground-background separation.
The datasets related to the above works,including FBMS \cite{ochs2014segmentation}, either avoid strongly camera translating sequences, or if it exists, provide a different ground-truth label of motion clusters (i.e. the background is not grouped as one cluster). These datasets are thus not appropriate for evaluating 3D motion segmentation approaches.}

\noindent\textbf{Multi-View Clustering} Multi-view learning aims to synergistically exploit the information derived from multiple modalities/views to achieve better learning. {The affinities induced by \textbf{H} and \textbf{F} here are seen as two different views. To avoid confusion with multi-frame motion segmentation, we use the term ``multi-model clustering/motion segmentation'' hereafter. } A detailed review on the current status and progress can be found in \cite{zhao2017multi}.
While extensive efforts have been dedicated to supervised multi-view/model learning \cite{Gonen2011}, that for the unsupervised case, in particular, clustering, is much less touched. 
We focus on the spectral framework for clustering \cite{von2007tutorial}, under which there are roughly two genres of multi-view approaches. The first kind discovers an optimal combination to aggregate multiple affinity matrices (kernels) for spectral clustering \cite{Huang2012,lange2006fusion,Wang2013}. However such combination is often non-trivial to discover. Alternatively, studies have been carried out on discovering a consensus on multiple kernels. In particular, the co-regularization scheme \cite{Kumar2011} was proposed to force data from different views to be close to each other in the embedding space for clustering. 
 Few if any of the existing approaches can guarantee superiority to the simple approach---kernel addition. In this work, we start our evaluation with this simplest baseline and then reveal its relation with the co-regularization schemes. {We also evaluate a custom-built version incorporating a subset constraint that preserves the true hierarchical structure of the affinity matrices induced by different geometric models}.

\noindent\textbf{Model Selection} This refers to the problem of estimating the number of clusters. 
We focus primarily on the model selection works in motion segmentation. One approach selects the {number of models} based on the multiplicities of the zero eigenvalue of the Laplacian matrix \cite{von2007tutorial}, or in the noisy case, using the gap heuristics \cite{Tibshirani2001}, the silhouette index \cite{rousseeuw1987silhouettes} or soft thresholding \cite{liu2013robust}.
As opposed to direct spectral analysis, another line of works have been developed for motion segmentation \cite{Li2013,Lai2017}. They first over-segment, i.e. estimate a higher number of clusters, and then a merging scheme is applied to obtain the final model parameter. These approaches achieved competitive performance on motion segmentation tasks; however, they lack a strong theoretical underpinning, e.g. the degree of over-segmentation is rather arbitrarily determined. The SCAMS approach \cite{Li2017} offers a principled characterization of the trade-off between various terms, specifically, a data fidelity term akin to our reconstruction error, and model complexity in terms of the rank and $L_0$ norm of the reconstructed affinity matrix \cite{Li2017}. Our work is similar in spirit in that we also have a global objective function, but the difference lies in that our data fidelity term is normalized, and instead of using rank and matrix norm to characterize the goodness of the clusters, we adopt the classical criteria of intra-cluster coherence and inter-cluster dissimilarity. These two well-known criteria introduced by Shi and Malik \cite{Shi2005} have been shown to be well-approximated by the rate of loss of relevance information \cite{raj2010information}, defined in the Information Bottleneck clustering method \cite{tishby2001data} as a representation of model complexity. {Model selection was also considered in the 2D motion segmentation work \cite{ochs2014segmentation} by exploiting a 2D spatial smoothness constraint. However, such assumption is often not true in 3D motion segmentation tasks.}


\section{Methodology}

In this section, we first describe the geometric models {(affine, homography and fundamental matrix)} used for motion segmentation and their hypothesis formation process. We then explain how the affinities between feature points are encapsulated in the ORK kernel\cite{Lai2017, Chin2009}. Finally, we explain the extension from single-model to multi-model clustering. In particular, we elaborate the relation between kernel addition and co-regularization for generic multi-kernel clustering, and we describe how the geometric relation that exists between models can be used to formulate a custom-made subset constrained multi-model clustering.

\subsection{Geometric Model Hypothesis}

Denote the observations of tracked points throughout $\scal{F}$ frames as $\{\vect{x}_{i}\}_{f=1\cdots F}$. We then randomly sample a minimal number of $p$ such points visible in a pair of frames and use them to fit a hypothesis of the model. The models tested include the fundamental matrix $\mathbf{F}$, homography $\mathbf{H}$, as well as the affine matrix $\mathbf{A}$. The reason for including the affine matrix model is because many existing datasets contain sequences with very weak perspective so this simpler model might be numerically more stable. For the three models $\mathbf{F}$, $\mathbf{H}$, and $\mathbf{A}$, the respective values for $p$ are 8, 4, and 3. \changed{The parameters of the model are estimated via direct linear transform (DLT) \cite{hartley2003multiple} and $500\times (F-1)$ hypotheses are sampled for each type of geometric model.}


\subsection{Affinity Captured as Ordered Residual Kernel}

Given multiple hypotheses $\{\matr{y}_k\}_{k=1\cdots K}$ generated from a particular model (affine, homography or fundamental matrix), we first compute for each data point the residual to all these hypotheses $\{d(\vect{x}_i,\matr{y}_k)\}_{k=1\cdots K}$ in terms of their Sampson errors \cite{hartley2003multiple}. The affinity between two features is captured in the correlation of preference for these hypotheses. Specifically, we can define the correlation in terms of the co-occurrence of points among all hypotheses. That is, if we define the indicator of point $\vect{x}_i$ being the inlier of all hypotheses $\{\matr{y}_k\}$ as $\vect{o}_i\in\{0,1\}^{\scal{K}}$, then the co-occurrence between two points is written as $k_{ij}=\vect{o}_i^\top\vect{o}_j$.
However, the threshold $\tau$ needed to determine when a data is an inlier (i.e. $\vect{o}_i=\mathbbm{1}(d(\vect{x}_i,\matr{H}_k)<\tau)$) is not easy to set, due to the potentially disparate range of motion present in different sequences.
The ordered residual kernel (ORK) \cite{Chin2009,Lai2017} was proposed to deal with this issue. Instead of fixing a threshold, the ORK sorts the residual in ascending order $\{\hat{d}_{i1} \quad \hat{d}_{i2} \cdots \hat{d}_{iK}\}$ where $\forall k : \hat{d}_{ik}<=\hat{d}_{ik+1}$. An adaptive threshold is then selected as the top $h$-th residual, i.e. $\tau_i=\hat{d}_{ih}$. The ORK kernel is also known to be resilient to serious sampling imbalance, an important advantage in real-world scenes where background is usually very large. Therefore, we adopt the ORK kernel to encapsulate the affinities between feature points. After constructing the affinity matrix, we normalize the affinities by dividing all $k_{ij}$ entries by the number of frames where both feature points $i$ and $j$ are visible. This step removes the weighting balance caused by incomplete trajectories.
Finally, as is customary in motion segmentation works, we subject the affinity matrix to a sparsification step. We use the $\epsilon$-neighborhood scheme of \cite{Lai2017} for this purpose.



\subsection{Spectral Clustering for Motion Segmentation}

We are now ready to use spectral clustering to recover the
clusters. We first review the single view spectral clustering
problem and then extend it to multi-view clustering.

\subsubsection{Single-Model Spectral Clustering}


Given the single affinity matrix $\matr{K}$, the normalized Laplacian $\matr{L}=\matr{I}-\matr{D}^{-0.5}\matr{K}\matr{D}^{-0.5}$ is first computed, where $\matr{D}$ is the degree matrix. The following objective is then set up to eigendecompose $\matr{L}$:

\begin{equation}
\min\limits_{\matr{U}} tr\left(\matr{U}^\top \matr{L} \matr{U}\right), \quad s.t. \matr{U}\matr{U}^\top=\matr{I}
\end{equation}
where $tr\left(\cdot\right)$ is the trace operator. The spectral embedding $\matr{U}\in\mathbbm{R}^{N\times M}$ can be efficiently solved and then treated as a new feature representation of the original points. A separate K-means step is then fed with the first $M$ dimensions of the normalized $\matr{U}$ for grouping points into $M$ motion groups.

\subsubsection{Multi-Model Spectral Clustering}

With multiple views provided by the different types of motion models, we have now at our disposal multiple affinity matrices. We explore two generic and one custom-made multi-model spectral clustering schemes to fuse the multiple sources of information together for clustering.

\noindent\textbf{Kernel Addition} A naive way to fuse information from heterogeneous sources for clustering is by kernel addition \cite{Kumar2011}. Given affinity matrices induced by heterogeneous sources $\{\matr{K}_v\}_{v=1\cdots V}$, kernel addition yields a fused kernel by summing up each individual kernel $\matr{K}=\sum_v \matr{K}_v$. With the corresponding Laplacian matrices written as $\matr{L}_v=\matr{I}-\matr{D}^{-0.5}_v\matr{K}_v\matr{D}^{-0.5}_v$, the objective for kernel addition can be written as,

\begin{equation}
  \resizebox{0.73\linewidth}{!}{$
\begin{split}
&\min\limits_{\matr{U}} tr(\vect{U}^\top\sum_v\matr{L}_v\vect{U}),\quad s.t. \vect{U}^\top\vect{U}=\matr{I}\\
\Rightarrow &\min\limits_{\{\matr{U}_v\}} \sum_v tr(\vect{U}_v^\top\matr{L}_v\vect{U}_v),\quad s.t. \vect{U}_v^\top\vect{U}_v=\matr{I}, \\
&\quad \forall v,w\in\{1,\cdots V\}: \vect{U}_v=\vect{U}_w
\end{split}$}
\end{equation}
\noindent
We note the kernel addition strategy is equivalent to discovering a common spectral embedding $\matr{U}$ among all models. This requirement of having a single consensus embedding can be too strong.

\noindent\textbf{Co-Regularization} Instead of demanding a common embedding, another solution is to include an additional regularization term in the objective function to encourage pairwise consensus between any two spectral embeddings $\matr{U}_v$ and $\matr{U}_w$. This has been studied by \cite{Kumar2011} who introduced a co-regularization term $tr\left(\matr{U}_v\matr{U}_v^\top\matr{U}_w\matr{U}_w^\top\right)$. This trace term returns high value if the new kernel matrix in the spectral embedding space $\matr{U}_v\matr{U}_v^\top$ and $\matr{U}_w\matr{U}_w^\top$ are similar to each other and vice versa. Incorporating the co-regularization term, we obtain the following objective:

\begin{equation}\label{eq:CoRegObj}
  \resizebox{0.8\linewidth}{!}{$
\begin{split}
\min_{\{\matr{U}_v\}} &\sum_v tr(\vect{U}_v^\top\matr{L}_v\vect{U}_v)-\lambda\sum_v\sum_wtr(\matr{U}_v\matr{U}_v^\top\matr{U}_w\matr{U}_w^\top),\\
s.t. &\vect{U}_v^\top\vect{U}_v=\matr{I}\\
\end{split}$}
\end{equation}

We can interpret the co-regularization scheme as a relaxed version of kernel addition. By increasing the penalty coefficient $\lambda$, the co-regularization scheme will approach kernel addition as all embeddings are forced to approach each other. This model is termed as pairwise co-regularization by \cite{Kumar2011} as the co-regularization term comprises of all pairs of spectral embeddings. The co-regularization model can be efficiently solved by initializing each view $\matr{U}_v$ separately in the same way as single-model spectral clustering. Then we recursively update each view with all other views fixed. When solving a single view, the problem becomes a standard eigendecomposition problem.
After convergence, we can concatenate the new spectral embedding of all views to produce an extended feature for the K-means step.

\subsubsection{Subset Constrained Multi-Model Spectral Clustering}

The above two multi-model spectral clustering schemes are generic fusion methods that do not exploit any relation that might exist between the different views. In the specific case of motion segmentation, we know that for any $\matr{H}$ between two frames, we can always define a family of $\matr{F}=[\vect{e}]_x\times \matr{H}$ parameterized by a vector $\vect{e}$, where $[\vect{e}]_x$ denotes the skew-symmetric matrix of $\vect{e}$ \cite{hartley2003multiple}. This means a pair of points that are the inliers of a homography should always be the inliers of a certain fundamental matrix. Conversely, if a pair of points are not the inliers of any $\matr{F}$, there is no homography which could take both points as inliers\footnote{We assume in the above two propositions that there are always enough points to fit an $\matr{F}$ if it exists.}. Generally speaking, we should expect that if $\mathbf{K_A}$, $\mathbf{K_H}$, and $\mathbf{K_F}$ are ideal binary affinity matrices, then $\mathbf{K_A} \le \mathbf{K_H} \le \mathbf{K_F}$, {where $\le$ indicates elementwise relation}. We term this hierarchical relationship the subset constraint. Imposing this constraint will help to further denoise or repair the affinity matrices. We cast this problem as a constrained clustering problem (adapted from \cite{Wang2014}):

\begin{equation}
\resizebox{.65\hsize}{!}
{$
\begin{split}
\min_{\{\matr{U}_v\}} &\sum_v tr\left(\matr{U}_v^\top\matr{L}_v\matr{U}_v\right) - \gamma tr(\matr{U}_v^\top\matr{Q}_v\matr{U}_v),\\
s.t. &\matr{U}_v^\top\matr{U}_v=\matr{I},\quad \matr{Q}_v\in \{-1,0,1\}^{N\times N}\\
\end{split}
$}
\end{equation}
\noindent
where the matrix $\matr{Q}_v$ provides the subset constraint for the $v$-th view. For $q_{ij}=1$, the constraint encourages a high inner product $\vect{u}_{vi}^\top\vect{u}_{vj}$ where $\vect{u}_{vi}$ indexes the $i$-th column. This means points $i$ and $j$ are encouraged to fall into the same cluster. For $q_{ij}=-1$, the constraint encourages a different cluster assignment between $i$ and $j$, and lastly, for $q_{ij}=0$, there is no constraint. For any single view $v$, the constraints $\matr{Q}_v$ is imposed by other views. For example, solving view $\matr{H}$, the positive constraint $q_{ij}$ is inherited from the result of $\matr{K_A}$; that is, if there is a link between points $i$ and $j$ from $\matr{K_A}$, then the $(i,j)$ entry of $\matr{K_H}$ is encouraged to be 1. On the other hand, the negative constraints come from $\matr{F}$. One could solve this problem using an alternating minimization scheme, but the subset constraint matrix $\matr{Q}_v$ may flip their values from 1 to -1 and vice versa in each alternating step, posing significant difficulties for convergence.

Therefore, we relax $\matr{Q}_v$ to continuous values. Instead of using the discretized results from other views, we use the affinity reconstructed from the spectral embedding $\matr{\hat{K}}=\matr{U}\matr{U}^\top$ to construct $\matr{Q}_v$ as detailed in Eq~(\ref{eq:Qconstraint}). We assume the three models are placed in the order of affine ($v=1$), homography ($v=2$) and fundamental matrix ($v=3$). The final objective is given by Eq~(\ref{eq:Qconstraint}).

\begin{equation}
\resizebox{1\hsize}{!}
{$
\begin{split}
&\min_{\{\matr{U}_v\}} \sum_v tr\left(\matr{U}_v^\top\matr{L}_v\matr{U}_v\right) - \gamma tr(\matr{U}_v^\top\matr{Q}_v\matr{U}_v),\quad s.t. \matr{U}_v^\top\matr{U}_v=\matr{I}, \\
&\matr{Q}_v = \begin{cases}
 \mathbbm{1}\left(\matr{\hat{K}}_{v+1} <0\right)\circ \matr{\hat{K}}_{v+1}, \quad v=1 \\
\mathbbm{1}\left(\matr{\hat{K}}_{v-1} >0\right)\circ\matr{\hat{K}}_{v-1} + \mathbbm{1}\left(\matr{\hat{K}}_{v+1} <0\right)\circ \matr{\hat{K}}_{v+1}, \quad v=2 \\
\mathbbm{1}\left(\matr{\hat{K}}_{v-1} >0\right)\circ \matr{\hat{K}}_{v-1}, \quad v=3
\end{cases}
\end{split}
$}\label{eq:Qconstraint}
\end{equation}
\noindent
where $\circ$ represents element-wise multiplication and $\mathbbm{1}\left(\cdot\right)$ is the indicator function. The subset constraint means for model $\matr{A}$ ($v=1$), only the negative constraint from $\matr{H}$ is applied, for model $\matr{H}$, both positive and negative constraints from $\matr{A}$ and $\matr{F}$ are applied respectively.  The final problem can be solved by optimizing each view $\matr{U}_v$ in an alternating fashion. We summarize the whole procedure in Algorithm~\ref{alg:Subset}.

\subsubsection{Convergence Analysis}

For both co-regularization and subset constrained clustering, we note the objective is not guaranteed to be convex w.r.t. all views' embeddings, depending on the values of $\lambda$ or $\gamma$. Nevertheless, we prove that the co-regularization model is guaranteed to converge to at least a local minimum. As we solve the problem in an alternating fashion, each step involves solving Eq~(\ref{eq:CoRegObj}) for the $v$-th view with all other views fixed as below,

\begin{equation}
\resizebox{.9\hsize}{!}
{$
\min\limits_{\matr{U}_v} tr\left(\matr{U}_v^\top\left(\matr{L}_v-\lambda\sum_{w,w\neq v}\matr{U}_w\matr{U}_w^\top\right)\matr{U}_v\right)\quad s.t. \matr{U}_v^\top\matr{U}_v=\matr{I}
$}
\end{equation}

The above problem can be efficiently solved by eigendecomposition regardless of the convexity of $\left(\matr{L}_v-\lambda\sum_{w,w\neq v}\matr{U}_w\matr{U}_w^\top\right)$. Therefore, solving all views iteratively results in a monotonically decreasing cost until convergence to a local minimum. The convergence for subset constrained clustering is, however, not guaranteed due to the constraint matrix $\matr{Q}_v$ changing at each iteration. Nevertheless, experimental results suggest that a proper selection of $\gamma$ say, less than $1e-2$, renders the problem amenable to convergence.


 \scalebox{0.85}{
    \begin{minipage}{1\linewidth}
 \removelatexerror
 \begin{algorithm}[H]
\SetKwData{Left}{left}\SetKwData{This}{this}\SetKwData{Up}{up}
\SetKwFunction{Concatenate}{Concatenate}\SetKwFunction{Kmeans}{K-means}
\SetKwInOut{Input}{input}\SetKwInOut{Output}{output}
\Input{\small{Kernel matrices $\{\matr{K}_v\}$, no. of motion $M$ and $\gamma$}}
\Output{Rigid motion index $\matr{X}$}
\emph{// Initialize Spectral Embedding}\\
\For{$v \leftarrow 1$ \KwTo $V$}
{
	Compute Laplacian matrix $\matr{L}_v=\matr{I}-\matr{D}^{-0.5}_v\matr{K}_v\matr{D}^{-0.5}_v$\;
	$\matr{U}_v \leftarrow$ first $M$ eigenvectors of $\matr{L}_v$\;
}
\emph{// Subset Constrained Spectral Clustering}\\
\While{Not Converged}
{
\For{$v \leftarrow 1$ \KwTo $V$}
{
	Compute $\matr{Q}_v$ following Eq~(\ref{eq:Qconstraint})\;
	Compute constrained Laplacian matrix $\tilde{\matr{L}}_v=\matr{L}_v - \gamma\matr{Q}_v$\;
	$\matr{U}_v \leftarrow$ first $M$ eigenvectors of $\tilde{\matr{L}}_v$\;
}}
\emph{// K-means to return index}\\
{$\matr{U}\leftarrow$ \Concatenate{$\matr{U}_1,\cdots\matr{U}_V$}
}\;
$\matr{X}\leftarrow$ \Kmeans{$\matr{U}$,$M$}
\caption{\small{Subset Constrained Clustering}\label{alg:Subset}}
\end{algorithm}
\end{minipage}}

\section{Model Selection}

In the preceding section, the final K-means step in the multi-model spectral clustering algorithm assumes that the number of clusters, i.e., the number of motion groups $M$ in Algorithm~\ref{alg:Subset} is known a priori. However, any motion segmentation algorithm, understood in the broadest term, should really estimate this parameter $M$ directly from the affinity matrix. This procedure is often termed as model selection in the clustering literature. In Section 1, we have proposed three criteria for model selection: data fidelity, intra-cluster coherence and inter-cluster dissimilarity. In this section, we introduce in greater details the various terms that represent the three criteria: the reconstruction error term for data fidelity, and the normalized cut term for the intra-cluster coherence and inter-cluster dissimilarity. Before doing so, we introduce some basic graph notation we will use for this discussion. Suppose the affinity matrix has been set up in a similarity graph $(\set{V},\set{E})$ where each vertex in $\set{V}$ represents a data point and each edge between two vertices the similarity between two data points. For a subset $\set{A}$ of $\set{V}$, $vol(\set{A})$ sums over the weights of all edges attached to vertices in $\set{A}$; for two disjoint subsets $\set{A}$ and $\set{B}$ of $\set{V}$, $cut(\set{A},\set{B})$ sums the edge weights between nodes in $\set{A}$ and $\set{B}$.

\subsection{Reconstruction Error}
Define an assignment matrix under a M-way clustering as,
\begin{equation}
\matr{X}=\left[\vect{x}_1 \cdots \vect{x}_M \right],\quad \vect{x}_i\in \{0,1\}^{N\times 1},\forall i\neq j:\vect{x}_i^\top\vect{x}_j=0
\end{equation}
\noindent
where $\vect{x}_i$ is the indicator vector for cluster $i$.
The corresponding $M$ clusters are denoted as $\{\set{A}_m\}_{m=1\cdots M}$. We can therefore reconstruct the ideal affinity matrix $\matr{W}$ as
\begin{equation}
 \resizebox{0.4\linewidth}{!}{$
\begin{split}
\matr{W}&=\matr{X}\matr{X}^\top=\sum_i \vect{x}_i\vect{x}_i^\top
\end{split}\label{eq:RecAff}$}
\end{equation}

For any feature pair $j,k$ in this reconstructed affinity matrix, $w_{jk}$ indicates if these two feature points belong to the same cluster and takes on only $\{0,1\}$ values. A good clustering should yield a $\matr{W}$ not too far from the original affinity matrix $\matr{K}$. We propose to measure such difference with the normalized Frobenius norm:

\begin{equation}
 \resizebox{0.6\linewidth}{!}{$
\begin{aligned}
\begin{split}
&\epsilon\left(\matr{W},\matr{K}\right) = \left\lVert\frac{\matr{X}\matr{X}^\top}{\left\lVert\matr{X}\matr{X}^\top\right\rVert_F} - \frac{\matr{K}}{\left\lVert\matr{K}\right\rVert_F}\right\rVert_F^2 \\
&= tr\left(\left(\frac{\matr{X}\matr{X}^\top}{\left\lVert\matr{X}\matr{X}^\top\right\rVert_F} - \frac{\matr{K}}{\left\lVert\matr{K}\right\rVert_F}\right)^\top\left(\frac{\matr{X}\matr{X}^\top}{\left\lVert\matr{X}\matr{X}^\top\right\rVert_F} - \frac{\matr{K}}{\left\lVert\matr{K}\right\rVert_F}\right)\right)\\
&= 2-\frac{2}{\left\lVert\matr{X}\matr{X}^\top\right\rVert_F  \left\lVert\matr{K}\right\rVert_F}tr\left(\matr{X}\matr{X}^\top\matr{K}\right)
\end{split}
\end{aligned}$}\label{eq:RecErr}
\end{equation}

It can be seen from the last line of the preceding equation that the reconstruction error is negatively related to the two matrices' Frobenius inner product $tr(\matr{X}\matr{X}^\top\matr{K})=\langle\matr{X}\matr{X}^\top,\matr{K}\rangle_F$ which characterizes the similarity between the two matrices. The reconstruction error is also positively related to the normalization factor $||\matr{X}\matr{X}^\top||_F$ which characterizes the sparsity of the reconstructed affinity (since $||\matr{X}\matr{X}^\top||_F^2=||\matr{X}\matr{X}^\top||_0$ for a binary matrix $\matr{X}$). From the graph-theoretic point of view, the inner product is equivalent to the following cut:
\begin{equation}
 \resizebox{0.6\linewidth}{!}{$
\begin{split}
tr\left(\matr{X}\matr{X}^\top\matr{K}\right)&=tr\left(\matr{X}^\top\matr{D}\matr{X}\right)-tr\left(\matr{X}^\top\matr{L}\matr{X}\right)\\
&=C-\sum_mcut\left(\set{A}_m,\bar{\set{A}_m}\right)
\end{split}$}
\end{equation}
{where $C=\sum_{i}d_{ii}$ is a constant. Therefore, as {$M$} increases, there will be more cut cost incurred. Furthermore, this is accompanied by a faster decrease in the normalization factor $||\matr{X}\matr{X}^\top||_F$, as a larger {$M$} would increase the sparsity. The overall effect is to reduce $\epsilon(\matr{W},\matr{K})$ with increasing {$M$}. Thus using the reconstruction error alone incurs the risk of over-estimating {$M$}.}

%

\subsection{Normalized Cut}
The two classical criteria of intra-cluster coherence and inter-cluster dissimilarity are simultaneously optimized by the normalized cut objective \cite{Shi2005}:
%
\begin{equation}
 \resizebox{0.7\linewidth}{!}{$
\begin{split}
Ncut&\left(\{\set{A}_m\}\right)=\sum_{m=1\cdots M}\frac{cut\left(\set{A}_m,\bar{\set{A}_m}\right)}{vol\left(\set{A}_m\right)}\\
&=\sum_{i=1\cdots M}\frac{\matr{x}_i^\top \matr{L}\matr{x}_i}{\vect{x}_i^\top \vect{x}_i}=tr\left(\left(\matr{X}^\top\matr{X}\right)^{-1}\matr{X}^\top\matr{L}\matr{X}\right)
\end{split}\label{eq:NCut}$}
\end{equation}


If the number of clusters $M$ is unknown and has to be estimated from minimizing the normalized cut cost, one can easily verify that a trivial solution would be obtained, i.e. $M=1$. Even if one ignores this trivial solution, there is still a tendency for the normalized cut to underestimate $M$. The reasoning is as follows: Given that the normalized Laplacian matrix $\matr{L}$ is positive-semidefinite, then the Rayleigh quotient \cite{horn1990matrix} satisfies $0\leq\lambda_{min}\leq \frac{\matr{x}_i^\top \matr{L}\matr{x}_i}{\vect{x}_i^\top \vect{x}_i}\leq \lambda_{max}$ where $\lambda_{min}$ and $\lambda_{max}$ are the minimal and maximal eigenvalues of $\matr{L}$. A large $M$ in the normalized cut objective would mean that there are more Rayleigh quotients, all of which are positive. This would likely lead to a higher cost for higher $M$, meaning that over-segmentation will be penalized. Although we cannot prove that this is always true as it depends on the specific data distribution and the corresponding partition configuration, under most data distribution, the normalized cut term does behave as a model complexity term. This viewpoint also concurs with the information-theoretic derivation of min-cut based clustering \cite{raj2010information}, which showed that in the well-mixed limit of graph diffusion, the normalized cut term can be well-approximated by the rate of loss of relevance information, defined in                          \cite{tishby2001data} as a representation of model complexity.


Our proposed combination of the normalized cut and the reconstruction error (NCRE) together controls the optimal selection of the cluster number, as it balances the trade-off between the cost of partitioning the graph (normalized cut) and keeping the partition as close to the original similarity graph as possible:

\begin{equation}
 \resizebox{0.7\linewidth}{!}{$
\begin{aligned}
\begin{split}
&\min_{\matr{X},M} Ncut\left(\{\set{A}_m\}\right) + \delta \epsilon\left(\matr{W},\matr{K}\right)\\
\Rightarrow &\min_{\matr{X},M} tr\left(\left(\matr{X}^\top\matr{X}\right)^{-1}\matr{X}^\top\matr{L}\matr{X}\right) - \delta \frac{2tr\left(\matr{X}\matr{X}^\top\matr{K}\right)}{\left\lVert\matr{X}\matr{X}^\top\right\rVert_F  \left\lVert\matr{K}\right\rVert_F}\\
& s.t. \quad \matr{X}\in \{0,1\}^{N\times M}, M\in \{1,\cdots N\}
\end{split}
\end{aligned}
$}\label{eq:Ncut_RecErr}
\end{equation}

Instead of optimizing Eq~(\ref{eq:Ncut_RecErr}), we directly search for $M$, since this number is likely to be small, and this considerably simplifies the problem. Firstly, a number of candidate partitions  are generated by varying $M$ from 1 to $N$ and these are denoted as $\{\matr{X}_M,M\}_{M=1\cdots N}$. Then, all candidates are evaluated against the cost in Eq~(\ref{eq:Ncut_RecErr}); the candidate with the minimal cost is selected as the best model. The multi-model extension is naturally formulated as the sum of costs from all views:
\begin{equation}
 \resizebox{0.75\linewidth}{!}{$
 \begin{split}
\min_{\matr{X},M} \sum_vtr\left(\left(\matr{X}^\top\matr{X}\right)^{-1}\matr{X}^\top\matr{L}_v\matr{X}\right) -
 \delta \sum_v \frac{2tr\left(\matr{X}\matr{X}^\top\matr{K}_v\right)}{\left\lVert\matr{X}\matr{X}^\top\right\rVert_F  \left\lVert\matr{K}_v\right\rVert_F}
 \end{split}\label{eq:MdlSelObj}$}
\end{equation}

An algorithm for choosing the optimal $M$ is summarized in Algorithm~\ref{alg:ModelSelect}.

\IncMargin{1em}
 \scalebox{0.85}{
    \begin{minipage}{1\linewidth}
 \removelatexerror
\begin{algorithm}[H]
\SetKwData{Left}{left}\SetKwData{This}{this}\SetKwData{Up}{up}
\SetKwFunction{Concatenate}{Concatenate}\SetKwFunction{Kmeans}{K-means}
\SetKwInOut{Input}{input}\SetKwInOut{Output}{output}
\Input{\small{Kernel matrices $\{\matr{K}_v\}$, and $\delta$}}
\Output{Optimal number of cluster $\hat{M}$}
\emph{// Initialize Spectral Embedding}\\
\For{$v \leftarrow 1$ \KwTo $V$}
{
	Compute Laplacian matrix $\matr{L}_v=\matr{I}-\matr{D}^{-0.5}_v\matr{K}_v\matr{D}^{-0.5}_v$\;
	$\matr{U}_v \leftarrow$ first $M$ eigenvectors of $\matr{L}_v$\;
}
\emph{// Single-Model or Multi-Model Spectral Clustering}\\
\For{$M \leftarrow 1$ \KwTo $N$}
{
{Get Assignment $\matr{X}_M$} from \textbf{Algorithm 1}\;
{Calculate Residual $r_M$ according to Eq~(\ref{eq:MdlSelObj})}\;
}
\emph{// Optimal No. of Cluster}\\
{$\hat{M}=arg\min_M r_M$}
\caption{\small{Estimate Number of Clusters}\label{alg:ModelSelect}}
\end{algorithm}
\end{minipage}
}
\DecMargin{1em}




\section{Experiment}

We first of all extensively evaluate existing motion segmentation and model selection approaches and our proposed methods on five motion segmentation datasets including the KT3DMoSeg benchmark proposed by us. The creation of KT3DMoSeg is then detailed in the following section. Finally, we present qualitative examples to analyze the success of our approach and the impact of individual models on motion segmentation.

\subsection{Dataset}
We carry out experiments on three extant motion segmentation benchmarks including the Hopkins155 \cite{Tron2007}, the Hopkins12 \cite{Rao2010} for testing incomplete trajectories, the MTPV62 \cite{Li2013} for testing stronger perspective effects and the newly created KT3DMoSeg \cite{Xu2018} for testing outdoor scenes with strong perspective effects and large camera motion.

\noindent\textbf{Hopkins 155} was the first large-scale dataset proposed for evaluating motion segmentation performance by Tron et al. \cite{Tron2007}.  It consists of 120 sequences with 2 motion groups and 35 sequences with 3 motion groups. It has been widely used by motion segmentation and subspace clustering works \cite{Elhamifar2013,Li2013,Magri2014,Li2015} and serves as the de facto standard benchmark for motion segmentation. \changed{The feature trajectories are provided by the dataset.}

\noindent\textbf{Hopkins 12} was created by \cite{Rao2010} based on \cite{Tron2007}. Three sequences are taken and re-combined to create 12 sequences with 2 to 3 motion groups. Within these sequences, 0 to 75 percent of the observation matrix entries are missing. \changed{The feature trajectories are provided by the dataset.}

\noindent\textbf{MTPV 62} was proposed for evaluating various real-world challenges in motion segmentation by \cite{Li2013}. This dataset consists of 50 clips from the Hopkins 155 dataset and 12 clips collected from real outdoor scenes. 9 clips contain strong perspective effects. \changed{The feature trajectories are provided by the dataset.}

\noindent\textbf{KT3DMoSeg} was created by \cite{Xu2018} by selecting 22 clips from the KITTI dataset \cite{Geiger2013IJRR}. The number of moving objects (including background) ranges from 2 to 5 and all sequences contain real-world effects such as strong perspectives and strong forward translations not seen in the traditional datasets.
We evaluate the performance on this dataset in terms of classification error \cite{Tron2007} for motion segmentation and correct rate for model selection. \changed{Feature trajectories are obtained with the method introduced in section~\ref{sect:GT}.}

\noindent\changed{\textbf{FBMS59} was created by \cite{ochs2014segmentation} for evaluating long-term motion segmentation performance. The dataset consists of training set and test set each with 29 and 30 sequences respectively.  The number of motions is kept unknown and estimated by model selection algorithms. \changed{Feature trajectories are obtained by dense tracking \cite{Sundaram2010} without further processing.}}

\noindent\changed{\textbf{ComplexBackground} was originally created for video segmentation \cite{narayana2012improvements,narayana2012background}. Our evaluation is based upon a variant of the dataset, termed the ComplexBackground motion segmentation dataset, in which new ground-truth masks on a subset of fives sequences have been provided for motion segmentation tasks by \cite{bideau2016detailed}. The number of motions is automatically estimated as with FBMS59. \changed{Feature trajectories are obtained by dense tracking \cite{Sundaram2010} without further processing.}}

\subsection{Motion Segmentation on Existing Benchmarks}

In this section, we extensively compare single-model and multi-model approaches on Hopkins155 benchmark \cite{Tron2007}. Specifically, for single-model, we evaluate using affine, homography and fundamental matrix as the single geometric model. For multi-model motion segmentation, we evaluated Kernel Addition (KerAdd), Co-Regularization (CoReg) \cite{Kumar2011} and {Subset Constrained Clustering (Subset)}. We fix the regularization parameter $\lambda$ and $\gamma$ at $10^{-2}$.
We also extensively compare with state-of-the-art approaches, including:  ALC \cite{Rao2010}, GPCA \cite{Vidal2008}, LSA \cite{Yan2006}, SSC \cite{Elhamifar2013}, TPV \cite{Li2013}, T-Linkage \cite{Magri2014}, S$^3$C \cite{Li2015}, RSIM \cite{Ji2016}, RV \cite{Jung2014a}, BB \cite{Hu2015} and MSSC\cite{Lai2017}. {Among all these, the ALC, GPCA, LSA, SSC and S$^3$C assume an affine camera projection. MSSC and TPV adopted homography and fundamental matrix respectively as the motion model.}
The results are presented in Table~\ref{tab:AllPerf}. For those algorithms which do not explicitly handle missing data, we recover the data matrix using Chen's matrix completion approach \cite{Chen2008}.

We make the following observations from the results. Firstly, with regards to the use of homography matrix as a single geometric model, our finding echoes the excellent results of earlier work such as MSSC \cite{Lai2017}. In fact, the simpler affine model has an even lower error figures. Clearly, the stitching argument (via virtual slices) put forth in Section 1 for explaining the success of homography applies to the affine case too, in particular under weak perspective views. For the fundamental matrix as a model, the performance is slightly worse-off. The reasons are manifold: strong camera rotation, limited depth relief, and not least the subspace overlap between different rigid motion groups, to which this richer fundamental matrix model is particularly susceptible.
Secondly, after fusing multiple kernels, we saw a boost in performance compared to single-model approaches, e.g. $0.36\%$ error for kernel addition and $0.31\%$ for subset constrained clustering on Hopkins155. Consistent boost in performance can be observed on Hopkins12 and MTPV62 as well. Usually, the fusion can produce the best of all performance regardless of the fusion scheme used. Even the simple kernel addition yields very competitive performance. This provides a strong option for real applications where parameter tuning is not desirable. 

\begin{table}[htbp]
\ra{0.9}
  \centering
  \caption{Motion segmentation results on Hopkins155, Hopkins12, MTPV62 and KT3DMoSeg datasets evaluated as classification error ($\%$). $^*$The best performing model (RPCA+ALC$_5$ is reported for ALC \cite{Rao2010}). $^{**}$ State-of-the-Art models' performances are reported for the sequences with correct number of moving objects. `$-$' cells indicate not reported or no public code is available.}
  \footnotesize
  \setlength\tabcolsep{1pt} 
  \resizebox{1\linewidth}{!}{
    \begin{tabular}{p{5em}p{1em}llllllllll}
    \toprule
    Models & \multicolumn{3}{c}{Hopkins155 \cite{Tron2007}} & \multicolumn{2}{c}{Hopkins12 \cite{Rao2010}} & \multicolumn{4}{c}{MTPV62 \cite{Li2013}$^{**}$} & \multicolumn{2}{c}{KT3DMoSeg} \\
    \cmidrule(lr){1-1}
    \cmidrule(lr){2-4}
\cmidrule(lr){5-6}
\cmidrule(lr){7-10} \cmidrule(lr){11-12}
    \multicolumn{1}{l}{\textit{\small{Existings}}} & \multicolumn{1}{c}{\footnotesize{2 Mot.}} & \multicolumn{1}{c}{\footnotesize{3 Mot.}} & \multicolumn{1}{c}{\footnotesize{All}} & \multicolumn{1}{c}{\footnotesize{Avg.}} & \multicolumn{1}{c}{\footnotesize{Med.}} & \multicolumn{1}{c}{\thead{Persp.\\ 9 clips}} & \multicolumn{1}{c}{\thead{Missing\\ 12 clips}} & \multicolumn{1}{c}{\thead{Hopkins\\ 50 clips}} & \multicolumn{1}{c}{\thead{All\\ 62 clips}} & \multicolumn{1}{c}{\footnotesize{Avg.}} & \multicolumn{1}{c}{\footnotesize{Med.}}\\
    \midrule
    LSA \cite{Yan2006} & 4.23  & 7.02  & 4.86  & -     & -     & -     & -     & -     & - & 38.30 & 38.58 \\
    GPCA \cite{Vidal2008} & 4.59  & 28.66 & 10.02 & -     & -     & 40.83 & 28.77 & 16.20 & 16.58 & 34.60 & 33.95\\
    {ALC \cite{Rao2010}} & 2.40  & 6.69  & 3.56  & 0.89$^*$ & 0.44$^*$ & 0.35  & 0.43  & 18.28 & 14.88 & 24.31 & 19.04\\
    SSC \cite{Elhamifar2013} & 1.52  & 4.40  & 2.18  & -     & -     & 9.68  & 17.22 & 2.01  & 5.17 & 33.88 & 33.54\\
    TPV \cite{Li2013} & 1.57  & 4.98  & 2.34  & -     & -     & 0.46  & 0.91  & 2.78  & 2.37 & - & -\\
    MSMC \cite{Dragon2012} & 3.04 & 8.76 & 4.33 & - & - & - & - & - & - & 27.74 & 35.80 \\
    LRR \cite{liu2013robust} & 1.33  & 4.98  & 1.59  & -     & -     & -  & -  & -  & - & 33.67 & 36.01\\
    \multicolumn{1}{l}{T-Lkg. \cite{Magri2014}} & 0.86  & 5.78  & 1.97  & -     & -     & -     & -     & -     & - & - & - \\
    S$^3$C  \cite{Li2015} & 1.94  & 4.92  & 2.61  & -     & -     & -     & -     & -     & - & - & - \\
    RSIM \cite{Ji2016} & 0.78 & 1.77 & 1.01 & 0.68 & 0.70 & - & - & - & - & - & - \\
    RV \cite{Jung2014a} & 0.44 & 1.88 & 0.77 & -     & -     & -     & -     & -     & - & - & - \\
    BB \cite{Hu2015} & - & - & 0.63 & -     & -     & -     & -     & -     & - & - & - \\
    \multicolumn{1}{l}{MSSC \cite{Lai2017}} & 0.54  & 1.84  & 0.83  & -     & -     & -     & 0.65  & \textbf{0.65}  & \textbf{0.65} & - & - \\
    \midrule
    \textit{Single-Model} & \multicolumn{11}{l}{} \\
    \midrule
    Affine & \textbf{0.40}  & \textbf{1.26}  & \textbf{0.59}  & {0.15}  & \textbf{0.10} &   \textbf{0.25}   & \textbf{0.35}  & \textbf{0.93}  & \textbf{0.82} & 15.76 & 11.52 \\
    Homo & 0.45  & 1.61  & 0.71  & \textbf{0.18}  & \textbf{0.10} &   0.70   & 0.48  & 1.23  & 1.08 & \textbf{11.45} & 7.14 \\
    Fund & 1.22  & 7.60  & 1.79  & 1.10  & \textbf{0.10} &   5.09  & 2.53  & 4.31  & 3.97 & 13.92 & \textbf{5.09} \\
    \midrule
    \textit{\small{Multi-Model}} & \multicolumn{11}{l}{} \\
    \midrule
    KerAdd & {0.27} & {0.66}  & 0.36  & {0.11} & \textbf{0.00} &  1.54   & 1.41  & 0.76  & 0.88 & 8.31 & 1.02 \\
    CoReg & 0.37  & {0.75} & {0.46} &   \textbf{0.06}  &  \textbf{0.00}     &   {0.22}   & \textbf{0.30} & {0.83} & {0.73} & \textbf{7.92} & {0.75} \\
     Subset & \textbf{0.23}  & \textbf{0.58} & \textbf{0.31} &   \textbf{0.06}  &  \textbf{0.00}     &   \textbf{0.20}   & \textbf{0.30} & {0.77} & \textbf{0.65} & {8.08} & \textbf{0.71} \\
    \bottomrule
    \end{tabular}%
    }
  \label{tab:AllPerf}%
\end{table}%

\subsection{Motion Segmentation on KITTI Benchmark}

The limitations of the Hopkins155 dataset are well-known: limited depth reliefs, dominant camera rotations, among others.
Such a dataset cannot meet the requirements of a benchmark meant for investigating motion segmentation capability in-the-wild, in particular, in self-driving scenario where the camera platform is often performing large translation and the scene is considerably more complex. For this reason, we propose a new motion segmentation benchmark based on the KITTI dataset \cite{Geiger2013IJRR}, the KITTI 3D Motion Segmentation Benchmark (KT3DMoSeg) \footnote{The dataset can be accessed from https://alex-xun-xu.github.io/ProjectPage/KT3DMoSeg/index.html}. We choose short video clips from the raw sequences of KITTI governed by three principles. Firstly, we wish to study sequences with more significant camera translation so camera mounted on moving cars are preferred. Secondly, we wish to investigate the impact of complex background structure; therefore, scene with strong perspective and rich clutter (in the structure sense) is selected. The moving objects could be of arbitrary (e.g. elongated) shapes rather than compact shapes like cubes or cones. Lastly, we are interested in the interplay of multiple motion groups, so clips with more than 3 motion groups are also chosen, as long as these moving objects contain enough features for forming motion hypotheses. In all, 22 short clips, each with 10-20 frames, are chosen for evaluation.

\subsubsection{Preprocessing and Labelling} \label{sect:GT}

In this section, we introduce the details of how we preprocess and label all clips.

\noindent\textbf{Tracking} First of all, we extract dense trajectories from each video clip using the code provided by \cite{Sundaram2010}. Key points are densely sampled on the whole frame with a gap of 8 pixels. Occlusion is detected by checking the consistency of the forward and backward flow \cite{Sundaram2010}. Trajectories which fail the check are considered to be occluded or the underlying flow is incorrect. These trajectories are stopped. In the next frame, new trajectories are densely sampled in those areas not occupied by existing trajectories. Finally, we further filter out short trajectories less than 5 frames for robustness. The resultant feature trajectories are very dense, on the order of 1500-5000 points per sequence, and more importantly, the background points account for an overwhelming majority of all feature points. 
This huge imbalance of background and foreground point sets renders hypothesis-driven methods liable to miss small foreground objects, and also impose high computational load on most algorithms. To relieve these problems, we sub-sample $10\%$ of the background points so that the average number of points is between 200-1000 for all sequences. The distribution of the number of points for all rigid motion groups is given in Fig.~\ref{fig:KTMoSeg_PerfPerSeq}(d).


%
%

\noindent\textbf{Manual Labelling} Due to the large number of tracked points, we need to come up with a method that can substantially reduce the effort in labelling. Our method is to only manually label the foreground moving objects, with the remaining unlabelled points all treated as stationary background points. Clearly, both the foreground and background obtained in this simple manner have many feature trajectories that do not belong well, either due to tracking errors or non-rigidities in the foreground motion. We next propose an efficient way to remove these outliers.

\noindent\textbf{Outlier Removal} We witness many erroneous trajectories generated by the dense tracking. Typical errors include those stemming from point drifting, in particular background points adhering themselves to moving foreground objects.
In this dataset, we identify outliers in a semi-autonomous manner with human-in-the-loop. In particular, we estimate via RANSAC a single fundamental matrix $\matr{F}$ over two frames using all points in each rigid motion group defined above. This is repeatedly done for all consecutive pairs of frames. The goodness of a trajectory is based on the sum of Sampson errors w.r.t the respective $\matr{F}$s along the trajectory and normalized by the number of frames the point has appeared in. Points with accumulated residuals greater than $Q3+7IQR$ are considered as outliers and removed; in the above, $Q3$ is the third quartile and $IQR$ is the inter-quartile range of all the residuals for points within a single motion. We do not claim that all bad features have been removed as a result, just as some small amount of bad feature trajectories still exist in the Hopkins 155 dataset. A completely automatic and reliable outlier detection module remains elusive, but addressing this problem is beyond the scope of our paper. To encourage further research on outlier detection for motion segmentation in-the-wild (i.e. without any manual intervention), we also publish the untrimmed feature trajectories. An illustration of the individual sequences with their various rigid motion groups and outliers is presented in Fig.~\ref{fig:GTtrajectories}, in which red dots indicate detected outliers.

\begin{figure*}
\begin{center}
\includegraphics[width=0.95\linewidth]{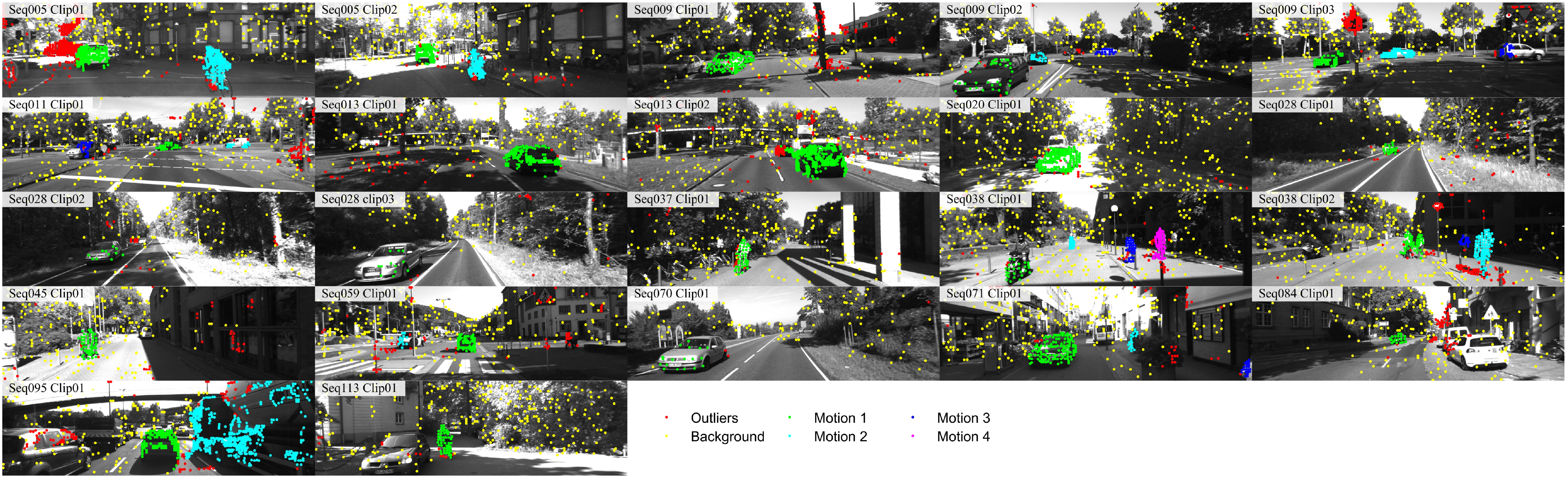}
\caption{Example frames of KT3DMoSeg dataset with trajectories overlapped. The color indicates different motion and outliers.}\label{fig:GTtrajectories}
\vspace{-0.5cm}
\end{center}
\end{figure*}


The same set of motion segmentation performance evaluation is carried out as in the preceding subsection and the results are presented in Table ~\ref{tab:AllPerf}. Both average and median classification errors are reported. The performances of the multi-model approaches are again consistently better than those of the single geometric model. Further evaluation on individual sequence is presented in Fig.~\ref{fig:KTMoSeg_PerfPerSeq} (a). To give some context to the performance figures, we use the ``Prevalence'' column to indicate the baseline solution of just assigning every feature as belonging to the prevalent group---the background. The overall performance of this baseline approach is $27.95\%$ which is pretty strong compared to many existing approaches. 
For the more recent and hypothesis-driven approach like MSSC, although we do not have the codes for evaluation, we can get an idea of its performance in KT3DMoSeg by looking at the result of homography model, due to its essential similarity to MSSC. Clearly, the homography model is able to replicate its strong performance ($11.45\%$) on this real-world dataset despite facing much stronger perspective effects. {The affine model performs the worst due to the strong perspective effect present in the dataset.} While all our single-model approaches turned in substantially better results than the baseline approach, it is also evident from the percentage errors that each single-model has difficulties in dealing with real-world effects. The various multi-model schemes, especially the co-regularization approach, can further improve the performance. {We noted the significant difference between the mean and median errors in the multi-model approaches. We attribute this to a few difficult sequences, e.g. ``Seq009\_Clip01'', ``Seq038\_Clip02'' and ``Seq095\_Clip01'', whose special configurations pose significant difficulties for 3D motion segmentation. Detailed analysis are made in Section~\ref{sect:Qualitative}.}



\begin{figure*}[ht]
\begin{center}
\subfloat[Error on individual sequence]{\includegraphics[width=0.45\linewidth]{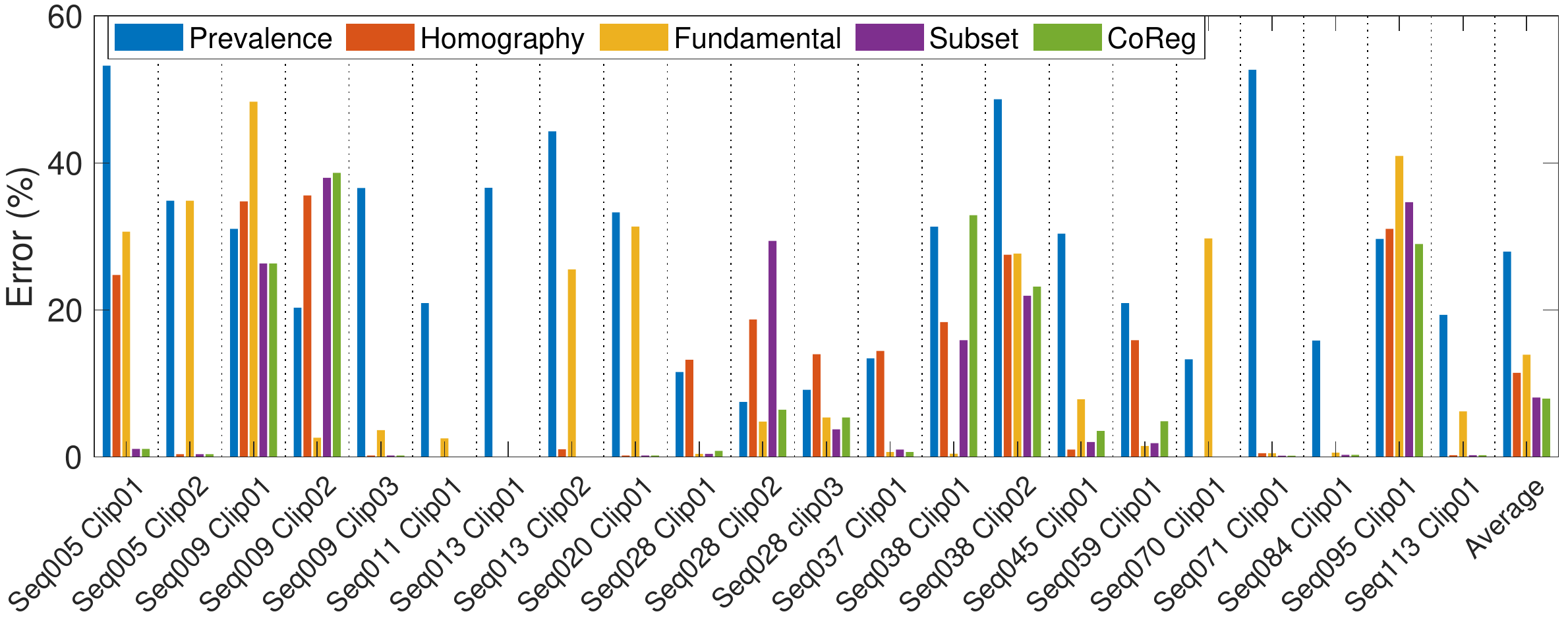}}
\subfloat[CoReg]{\includegraphics[width=0.135\linewidth]{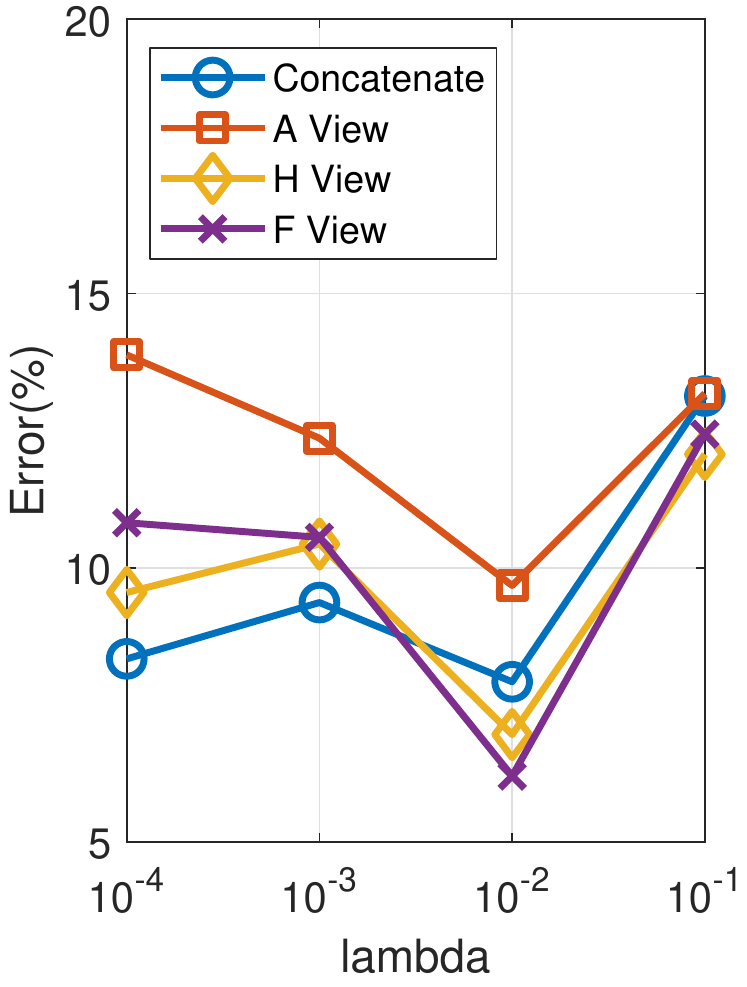}}
\subfloat[SubSet]{\includegraphics[width=0.135\linewidth]{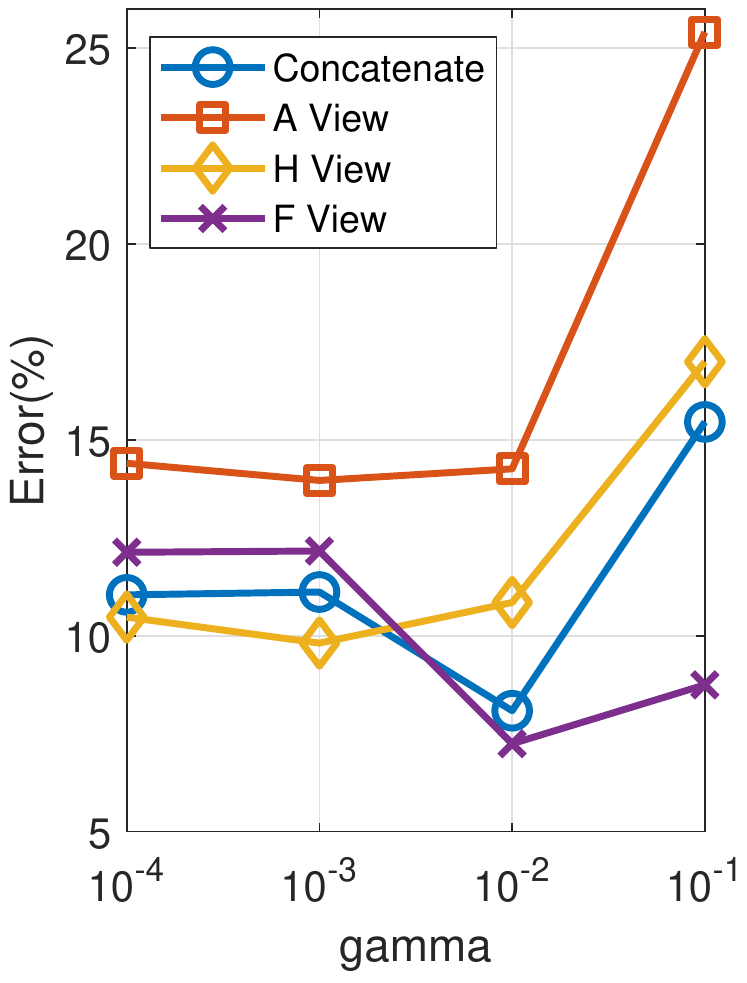}}
\subfloat[Distribution of Points]{\includegraphics[width=0.27\linewidth]{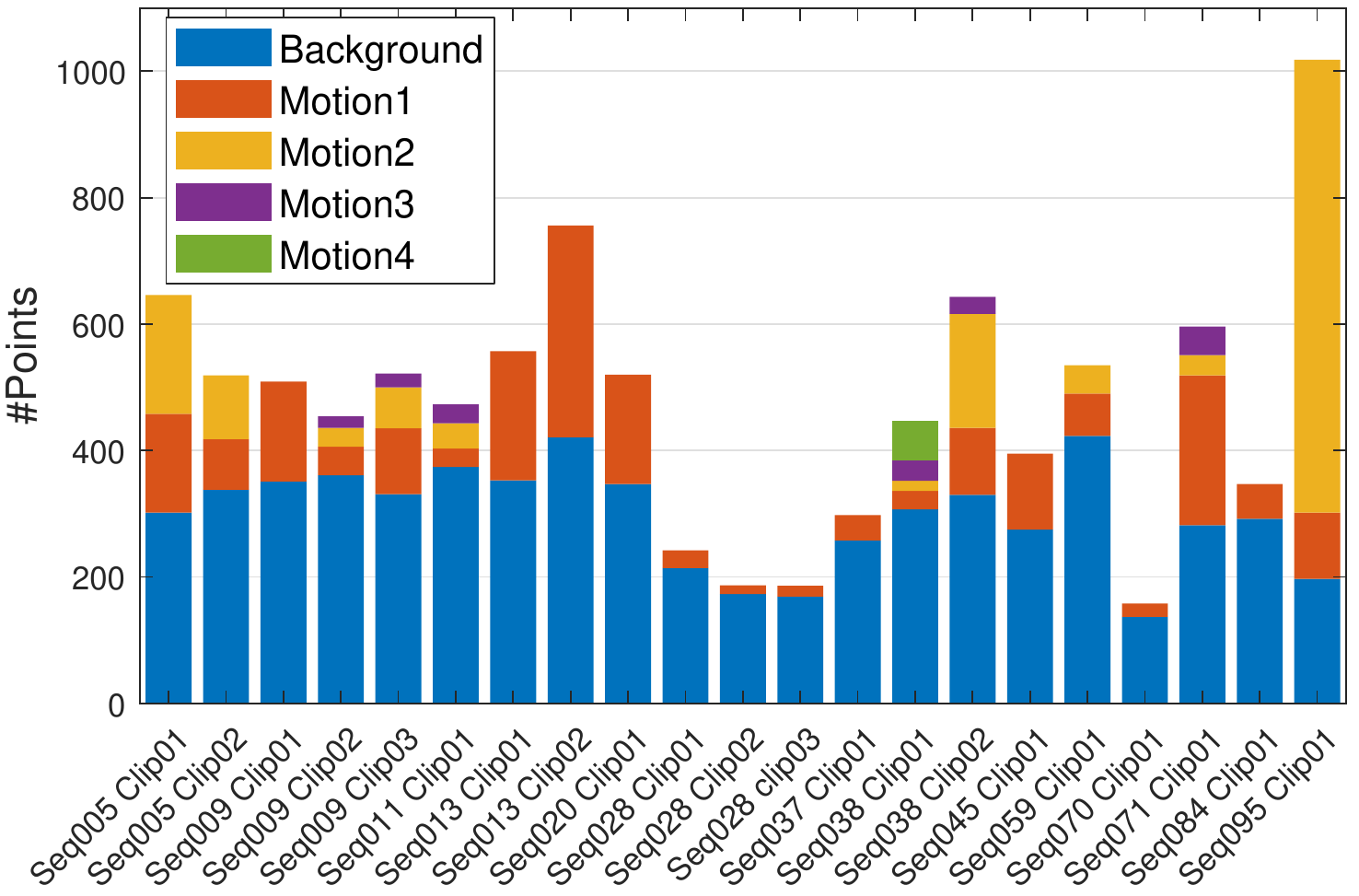}}
\vspace{-0.3cm}
\caption{ (a) The classification error on individual sequence of KT3DMoSeg dataset. (b-c) The impact of regularization parameters on co-regularization and subset constrained clustering performances. (d) The distribution of point number for each sequence.}
\vspace{-0.6cm}
\label{fig:KTMoSeg_PerfPerSeq}
\end{center}
\end{figure*}


\subsection{Model Selection}

In this section, we extensively evaluate the performance of our NCRE algorithm on model selection in terms of both the classification error and correct rate, i.e. the number of sequences with correctly estimated number of moving objects. This evaluation is carried out over three motion segmentation benchmarks, Hopkins 155, MTPV62 and KT3DMoSeg, and under both the single-model and multi-model settings. {The $\delta$ in Eq~(\ref{eq:Ncut_RecErr}) is specifically set to 0.1 for Hopkins 155, MTPV 62 and Hopkins 12 and 1 for KT3DMoSeg}. For all three datasets, we implement simple model selection approaches for comparison, including Self-Tuning Spectral Clustering \cite{zelnik2005self} and {Gap Heuristic, i.e. $M = \arg\max_m \left(\sigma_{m+1}-\sigma_m\right)$}; in both these approaches, the affinity matrix is provided by the homography model. We also extensively compare with state-of-the-art model selection approaches for motion segmentation. Specifically, we present the results of ALC \cite{Rao2010}, GPCA \cite{Vidal2008}, LSA \cite{Yan2006}, ORK \cite{Chin2009}, KO \cite{Chin2010}, LBF \cite{Zhang2012}, SSC \cite{Elhamifar2013}, MTPV \cite{Li2013}, LRR \cite{liu2013robust}, MB-FLoSS \cite{Lee2013}, BB \cite{Hu2015} and MSSC\cite{Lai2017} on Hopkins 155 and MTPV62. The GPCA, LBF and SSC are implemented with the second order difference (SOD)\cite{Zhang2012} method for selecting the optimal number of moving objects. It is worth noting that GPCA, LBF, SSC, self-tuning and its variant MSSC exclude the possibility of a single cluster. Since the sequences tested indeed do not contain the case of single rigid motion, advantages over other competitors are gained by these approaches. To allow fair comparison, we evaluate the performance of our NCRE algorithm under two cases: when the candidate number of cluster is allowed to range {from 1 to $M_{max}$, and from 2 to $M_{max}$. The latter permits comparison with approaches such as GPCA, self-tuning, MSSC, etc. which make similar assumption, while the former permits comparison with all others. We fix $M_{max}=10$ for all experiments. Further increasing $M_{max}$ would not affect the model selection accuracy but instead increase computation cost.}

We present the model selection results on Hopkins155 in Table~\ref{tab:Hopkins155MdlSel}, in which the performances of our proposed method under the two different ranges of candidate number of clusters are separated by a slash. It is obvious that under single-model affinity matrix, our model selection criteria are able to identify the number of moving objects with higher accuracy than all existing models with single-model affinity matrix. In particular, the homography model yields the best performance of all, $87.10\%$ for {1 to $M_{max}$} candidate no. of motion groups and $92.26\%$ for {2 to $M_{max}$} candidate no. of motion groups. This leads to an improvement of more than $6\%$ over the state-of-the-art result, MSSC with $85.81\%$. It corroborates our previous analysis that since the Hopkins155 dataset is rotation-dominant, homography is the most suitable geometric model. {The advantage in the model selection problem is clearer because, without knowing the number of motion groups a priori, the algorithm has to rely more on the inherent quality of the modelling. The effect of errors in other models, in particular \textbf{F}, are also felt more, limiting the benefits that can be reaped by the fusion method compared to the best single-model method. By fusing three models, we observe the correct rate in model selection to be on par with the best single-model (homography), while the classification error experiences a significant drop from $3.95\%$ and $2.70\%$ to $3.17\%$ and $1.83\%$ respectively. More detailed investigations into the impact of individual models are made in Section~\ref{sect:IndividualViews}.}

\begin{table}[!htbp]
  \centering
\caption{Model selection results on Hopkins 155 dataset. Both mean classification error ($\%$) and correct rate ($\%$) are reported for comparison. $^*$ indicates the candidate no. of motion groups is from 2 to $N$. `$-$' cells indicate not reported or no public code is available.}\vspace{-0.2cm}
  \footnotesize
  \setlength\tabcolsep{1pt} 
  \resizebox{1\linewidth}{!}{    \begin{tabular}{p{1em}p{5.8em}llllll}
    \toprule
    \multirow{10}[4]{*}{\begin{sideways}Existing Approaches\end{sideways}} & \multicolumn{1}{c}{\multirow{2}[2]{*}{Method}} & \multicolumn{2}{c}{2 Motion} & \multicolumn{2}{c}{3 Motion} & \multicolumn{2}{c}{All Seqs} \\
    \cmidrule(lr){3-4} \cmidrule(lr){5-6} \cmidrule(lr){7-8}
          &       & Error & CorrectRate & Error & CorrectRate & Error & CorrectRate \\
\cmidrule{2-8}          & ORK\cite{Chin2009} & 7.83  & 67.37 & 12.62 & 49.66 & 8.91  & 63.37 \\
          & KO\cite{Chin2010} & -     & 82.50 & -     & 48.57 & -     & 74.84 \\
          & LRR\cite{liu2013robust} & 8.59  & 84.17 & 15.51 & 57.14 & 10.16 & 78.06 \\
          & MB-FLS\cite{Lee2013} & 9.54  & 81.67 & 12.07 & \textbf{71.43} & 10.04 & 79.35 \\
          & BB\cite{Hu2015} & -     & -     & -     & -     & 6.09  & 80.00 \\
          & GapHeu & 8.25  & 84.17 & 10.06 & 54.29 & 8.66  & 77.42 \\
          & S.T. \cite{zelnik2005self} & 3.60$^*$  & 89.17$^*$ & 9.88$^*$  & 54.29$^*$ & 5.01$^*$  & 81.29$^*$ \\
          & MSSC\cite{Lai2017} & \textbf{2.50}$^*$ & \textbf{90.00}$^*$ & \textbf{7.15}$^*$ & \textbf{71.43}$^*$ & \textbf{3.55}$^*$ & \textbf{85.81}$^*$ \\
    \midrule
    \multirow{8}[8]{*}{\begin{sideways}Our Approaches\end{sideways}} & \multicolumn{7}{l}{\textit{Single-Model}} \\
\cmidrule{2-8}          & Affine & 4.34/{2.13}$^*$ & 87.50/96.67$^*$ & 6.37/6.23$^*$ & 65.71/65.71$^*$ & 4.80/3.06$^*$ & 82.58/89.68$^*$ \\
          & Homo  & \textbf{3.42}/2.14$^*$ & \textbf{90.00}/95.83$^*$ & \textbf{5.76}/\textbf{4.60}$^*$ & \textbf{77.14}/\textbf{80.00}$^*$ & \textbf{3.95}/\textbf{2.70}$^*$ & \textbf{87.10}/\textbf{92.26}$^*$ \\
          & Fund  & 9.90/3.40$^*$ & 72.50/\textbf{97.50}$^*$ & 11.7/10.85$^*$ & 42.86/45.71$^*$ & 12.31/5.08$^*$ & 65.81/85.81$^*$ \\
\cmidrule{2-8}          & \multicolumn{7}{l}{\textit{Multi-Model}} \\
\cmidrule{2-8}          & KerAdd & \textbf{2.31}/2.19$^*$ & \textbf{94.17}/97.50$^*$ & 7.62/7.71$^*$ & 57.14/60.00$^*$ & 3.51/3.44$^*$ & 85.81/89.03$^*$ \\
          & CoReg & 2.90/2.35$^*$ & 90.83/95.83$^*$ & \textbf{4.09}/\textbf{4.09}$^*$ & \textbf{74.29}/\textbf{74.29}$^*$ & \textbf{3.17}/2.74$^*$ & \textbf{87.10}/90.32$^*$ \\
          & Subset & 2.98/\textbf{0.43}$^*$ & 90.83/\textbf{100.00}$^*$ & 4.77/6.65$^*$ & 71.43/62.86$^*$ & 3.38/\textbf{1.83}$^*$ & 86.45/\textbf{91.61}$^*$ \\
    \bottomrule
    \end{tabular}%
    }
  \label{tab:Hopkins155MdlSel}%
  \vspace{-0.3cm}
\end{table}%

We further present the results on MTPV62 in Table~\ref{tab:MdlSel3Datasets}, in which we again present both the results with {1 to $M_{max}$} candidate no. of moving groups and {2 to $M_{max}$} candidate no. of moving groups. We make the following observations from the results. Both our single-model and multi-model approaches outperform the state-of-the-arts with large margin. With {2 to $M_{max}$} candidate, we improve the classification error from $5.09\%$ of MSSC to $2.78\%$ (Affine model) and $3.23\%$ (multi-model, with Subset constraint clustering). The number of sequence with correctly estimated cluster also increased to 53 from the previous best of 49 of MSSC. {We show the number of sequence that is correct rather than the percentage for model selection evaluation here so as to be consistent with past works on this benchmark.} { 
All our multi-model clustering variants produce the best performance of all in model selection and segmentation. It is worth noting that the gap between ALC and ours is the smallest among all other datasets. This is attributed to the severe degeneracy of Hopkins12 which is dominated by pure camera rotation, rendering the affine camera model in ALC adequate.}

\begin{table*}[htbp]
 \begin{minipage}{0.76\textwidth}
  \centering
  \caption{\small{Model selection performance on MTPV62, KT3DMoSeg and Hopkins12. The methods and results with $^*$ are based on the candidate no. of clusters from 2 to $M_{max}$. \changed{Both mean error (MeanErr) and correct rate (CorrectRate) are measured in $\%$. The correct number of sequence (Corr. \#Seq) counts the sequences with correctly estimated number of motion.}
  }}
  \footnotesize
   \setlength\tabcolsep{1pt} 
   \resizebox{1\linewidth}{!}{
   {
    \begin{tabular}{llllllllll}
\toprule          & \multicolumn{5}{c}{MTPV62}            & \multicolumn{2}{c}{KT3DMoSeg} & \multicolumn{2}{c}{Hopkins12} \\
\cmidrule(lr){2-6}\cmidrule(lr){7-8}\cmidrule(lr){9-10}          &       & \multicolumn{3}{c}{\changed{MeanErr}} &       & \multicolumn{1}{p{3.8em}}{\multirow{2}[4]{*}{\changed{MeanErr}}} & \multicolumn{1}{p{6em}}{\multirow{2}[4]{*}{\changed{CorrectRate}}} & \multicolumn{1}{p{3.8em}}{\multirow{2}[4]{*}{\changed{MeanErr}}} & \multicolumn{1}{p{3.8em}}{\multirow{2}[4]{*}{\changed{CorrectRate}}} \\
\cmidrule{2-5}    \multicolumn{1}{c}{Method} & \multicolumn{1}{p{4.5em}}{Persp.\newline{}9 Clips} & \multicolumn{1}{p{4.8em}}{Missing\newline{}12 Clips} & \multicolumn{1}{p{4.8em}}{Hopkins\newline{}50 Clips} & \multicolumn{1}{p{4.8em}}{All \newline{}62Clips} & \multicolumn{1}{p{3.8em}}{Corr. \newline{}\#Seq} &       &       &       &  \\
    \midrule
    ALC \cite{Rao2010} & 16.18 & 25.38 & 22.03 & 22.67 & \multicolumn{1}{l}{21} & 34.72 & 45.45 & 4.33  & 91.67 \\
    GPCA\cite{Vidal2008} & 43.66* & 39.64* & 16.89* & 21.29* & \multicolumn{1}{l}{33*} & 47.35* & 18.18* &       &  \\
    LBF \cite{Zhang2012} & 20.00* & 20.17* & 15.66* & 16.53* & \multicolumn{1}{l}{29*} & 62.86* & 18.18* & 32.63* & 33.33* \\
    LRR \cite{liu2013robust} & 16.31 & 26.03 & 9.82  & 12.98 & \multicolumn{1}{l}{35} & 26.94 & 31.82 & 18.86 & 41.67 \\
    MSMC \cite{Dragon2012} & 19.17 & 14.64 & 14.19 & 14.27 & \multicolumn{1}{l}{25} & 50.17 & 18.18 & 24.41 & 33.33 \\
    ORK \cite{Chin2009} & 22.94 & 24.11 & 12.98 & 15.13 & \multicolumn{1}{l}{37} & -     & -     & -     & - \\
    SSC \cite{Elhamifar2013} & 26.58* & 27.41* & 13.09* & 15.86* & \multicolumn{1}{l}{33*} & 64.82* & 18.18* & 49.18* & 25.00* \\
    MTPV \cite{Li2013} & 8.20  & 7.71  & 7.56  & 7.59  & \multicolumn{1}{l}{46} & -     & -     & -     & - \\
    MSSC\cite{Lai2017} & {-} & \textbf{1.84*} & \textbf{5.87*} & \textbf{5.09*} & \multicolumn{1}{l}{\textbf{49*}} & -     & -     & -     & - \\
    S.T. \cite{zelnik2005self} & 7.10* & 3.12* & 8.11* & 7.15* & \multicolumn{1}{l}{44*} & \textbf{22.01*} & \textbf{50.00*} & \textbf{4.05*} & \textbf{91.67*} \\
    GapHeu & -     & -     & -     & -     & \multicolumn{1}{l}{-} & 31.22 & 27.27 & 6.16  & \textbf{91.67} \\
    \midrule
    \multicolumn{10}{l}{\textit{Single-Model}} \\
    \midrule
    Affine & 2.68/1.84* & 1.28/0.33* & \textbf{3.59}/\textbf{3.37*} & \textbf{3.14}/\textbf{2.78*} & \textbf{50}/\textbf{53*} & \textbf{24.99}/19.08* & 36.36/54.55* & \textbf{3.91}/\textbf{2.82*} & \textbf{83.33}/\textbf{91.67*} \\
    Homography & \textbf{1.76}/\textbf{0.77*} & \textbf{1.25}/\textbf{0.24*} & 5.08/4.40* & 4.34/3.59* & 48/52* & 29.28/24.63* & 50.00/54.55* & 6.66/3.55* & 75.00/83.33* \\
    Fundamental & 12.23/7.65* & 20.49/5.19* & 13.46/9.22* & 14.82/8.92* & 23/43* & 28.22/\textbf{10.31*} & \textbf{54.55/63.64*} & 8.86/4.73* & 75.00/83.33* \\
    \midrule
    \multicolumn{10}{l}{\textit{Multi-Model}} \\
    \midrule
    KerAdd & \textbf{2.30}/0.63* & \textbf{1.85}/\textbf{0.26*} & 5.13/5.16* & 4.49/4.21* & \textbf{48}/\textbf{53*} & \textbf{10.21}/\textbf{7.51*} & \textbf{50.00/68.18*} & 6.63/\textbf{1.51*} & 75.00/\textbf{91.67*} \\
    CoReg & 4.10/\textbf{0.34*} & 3.37/0.41* & 3.93/4.45* & 3.82/3.67* & \textbf{48}/52* & 14.61/11.33* & \textbf{50.00}/63.64* & 5.91/2.26* & 83.33/\textbf{91.67*} \\
    Subset & 4.10/4.96* & 3.28/3.60* & \textbf{3.93/3.14*} & \textbf{3.81}/\textbf{3.23*} & 45/52* & 11.90/14.74* & \textbf{50.00/68.18*} & \textbf{5.46}/4.47* & \textbf{91.67/91.67*} \\
    \bottomrule
    \end{tabular}%
    }
    }
  \label{tab:MdlSel3Datasets}%
  \end{minipage}
    \hspace{-0.2cm}
    \changed{
   \begin{minipage}{0.24\textwidth}
   \centering
  \caption{\small{Comparisons on the first 10-frame subset of FBMS59 and the ComplexBackground dataset. All measures are in $\%$. }}
  \setlength\tabcolsep{3pt} 
	  \resizebox{1\linewidth}{!}{
    \begin{tabular}{lllll}
    \toprule
    \multicolumn{5}{c}{FBMS59 - TrainingSet} \\
    \midrule
    Model & Dens. & Prec. & Rec.  & F-m. \\
    \midrule
    ALC\cite{Rao2010}   & 0.12  & 54.31 & 54.80 & 54.56 \\
    SSC\cite{Elhamifar2013}   & 0.89  & 80.59 & 63.32 & 67.70 \\
    Ochs\cite{ochs2014segmentation}  & 0.95  & \textbf{92.77} & 65.44 & 76.75 \\
    MC\cite{keuper2015motion}    & \textbf{1.01} & 91.09 & 67.78 & 77.72 \\
    Subset(Ours) & 0.89  & 90.35 & \textbf{73.57} & \textbf{81.10} \\
    \midrule
    \multicolumn{5}{c}{FBMS59 - TestSet} \\
    \midrule
    Model & Dens. & Prec. & Rec.  & F-m. \\
    \midrule
    ALC\cite{Rao2010}   & 0.12  & 53.11 & 56.40 & 54.70 \\
    SSC\cite{Elhamifar2013}   & 0.88  & \textbf{91.67} & 50.57 & 65.18 \\
    Ochs\cite{ochs2014segmentation}  & 0.97  & 87.44 & 60.77 & 71.71 \\
    MC\cite{keuper2015motion}    & \textbf{1.01} & 89.05 & 61.81 & 72.97 \\
    Subset(Ours) & 0.90  & 84.41 & \textbf{72.87} & \textbf{78.22} \\
    \midrule
    \multicolumn{5}{c}{ComplexBackground} \\
    \midrule
    Model & Dens. & Prec. & Rec.  & F-m. \\
    \midrule
    Ochs\cite{ochs2014segmentation}  & 0.86  & 83.77 & 48.76 & 61.64 \\
    MC\cite{keuper2015motion} & \textbf{0.90} & \textbf{93.54} & 64.28 & \textbf{76.20} \\
    Subset(Ours) & \textbf{0.90} & 90.03 & 59.67 & 71.77 \\
    \bottomrule
    \end{tabular}%
    }
  \label{tab:FBMS59}%
   \end{minipage}
   }
   \vspace{-0.5cm}
\end{table*}%

We finally present the model selection results on KT3DMoSeg dataset in Table~\ref{tab:MdlSel3Datasets}. We implement multiple existing approaches on model selection including ALC\cite{Rao2010}, GPCA \cite{Vidal2008}, LBF\cite{Zhang2012}, LRR\cite{liu2013robust}, MSMC\cite{Dragon2012}, SSC\cite{Elhamifar2013} and Self-Tuning spectral clustering \cite{zelnik2005self}. We draw similar conclusions from the results as above. Both the single-model and multi-model approaches outperform the existing ones in both mean classification error and estimating the number of clusters. The significant perspective effect in this dataset renders the methods with affine projection assumption ineffective.

\subsection{Dense and Articulated Motion Segmentation}

\changed{We evaluate the performance on FBMS59\cite{ochs2014segmentation} dataset which was designed for dense motion segmentation. It is worth noting that many sequences of this dataset are captured by hand-held cameras with vanishing camera translation. Many sequences also involve moderate articulated motion. Thus the challenge of this dataset mainly lie in handling densely tracked feature points and articulation. We adopt evaluation on the first 10 frame subset, as proposed by \cite{ochs2014segmentation}, for fair comparison with traditional methods. The precision (Prec.), recall (Rec.) and F-measure (F-m) are evaluated as the metric and the density (Dens.) is measured as the coverage of all labelled pixels, all of which are higher the better. The feature trajectories are extracted by dense tracking\cite{Sundaram2010} with the default 8 pixels gap. We compare with ALC\cite{Rao2010}, SSC\cite{Elhamifar2013}, SpectralClustering (Ochs)\cite{ochs2014segmentation} and MultiCut(MC)\cite{keuper2015motion}. The hyperparameters of our method is searched over the training set and fixed for test set. The results are presented in Tab.~\ref{tab:FBMS59}. We observe very competitive performance of our final fusion model compared against the existing ones based on translational models and the improvements is significant on both the training and test set with particular improvement on the recall measure.\\
We further evaluate our approach on the ComplexBackground dataset \cite{bideau2016detailed}. We notice that there are many extremely small objects labelled, e.g. the vehicles in the faraway background of sequence `traffic' in Fig.~\ref{fig:ComplexBackground}, posing great challenge to existing feature trajectory based motion segmentation algorithms. For comparison, we adopted the same evaluation metrics as for FBMS59 and compared with Ochs\cite{ochs2014segmentation} and MC\cite{keuper2015motion}. The results are presented in Tab.~\ref{tab:FBMS59}. We observe superior performance of our algorithm compared with \cite{ochs2014segmentation}. The gap between MultiCut and ours is mainly due to the ground-truth definition favoring over-segmentation. We shall analyze in more details in the following section.
}

\subsection{Further Analysis}

We give insight into the success of our proposed multi-model motion segmentation and model selection methods and then analyze the impact of individual models.

\subsubsection{Qualitative Study}\label{sect:Qualitative}

\noindent\textbf{Motion Segmentation}
We now present the motion segmentation results on some sequences from KT3DMoSeg in Fig.~\ref{fig:QualitativeKT3DMoSeg} to better understand how different geometric models complement each other, as well as to illustrate the challenges posed by this dataset. All these sequences involve strong perspective effects in the background but the foreground moving objects often have limited depth reliefs and camera is making significant translation. Many background objects have non-compact shapes, and thus the slicing effect induced by the homography/affine model is less likely to relate all the background points together due to the lower connectivity. Therefore the background tends to split in the affine and homography models, e.g. the traffic sign in Fig.~\ref{fig:QualitativeKT3DMoSeg} (a), the wall in (e), the bush in (f) and the ground in (g). {While fundamental matrix is more likely to discover a seamless background in theory, it is plagued by a greater susceptibility to subspace overlap in practice. 
In (g), the fundamental matrix model identifies the whole background; however it fails to separate the white SUV from the background. In both (a), (e) and (g), the fusion schemes manage to correct these errors. There are also some challenges that remain in this dataset. Clearly, when the motion of the foreground object (e.g. the person in the middle of Fig.~\ref{fig:QualitativeKT3DMoSeg} (c), indicated by red points in GroundTruth) is small or intermittent compared to that of the camera, it can be difficult to detect. Coupled with the large depth range in the background, the algorithm can be fooled to split the background instead of segmenting the foreground. Lastly, scenes like Fig.~\ref{fig:QualitativeKT3DMoSeg} (b) and (d)  still pose serious challenge. It is well known that the epipolar constraint allows a freedom to translate along the epipolar line. This allows an independent motion that is moving with respect to the background but consistent with the epipolar constraint to go undetected.} In the figure (d), the car in front can be interpreted as a background object on the horizon, and thus the algorithm ends up splitting the big truck instead. The fusion scheme occasionally fails, as in (f) and (h), where the incorrect split in affine/fundamental matrix model is propagated to the final result and cause unexpected segmentation results.

\begin{figure*}[!h]
\begin{center}
\includegraphics[width=0.98\linewidth]{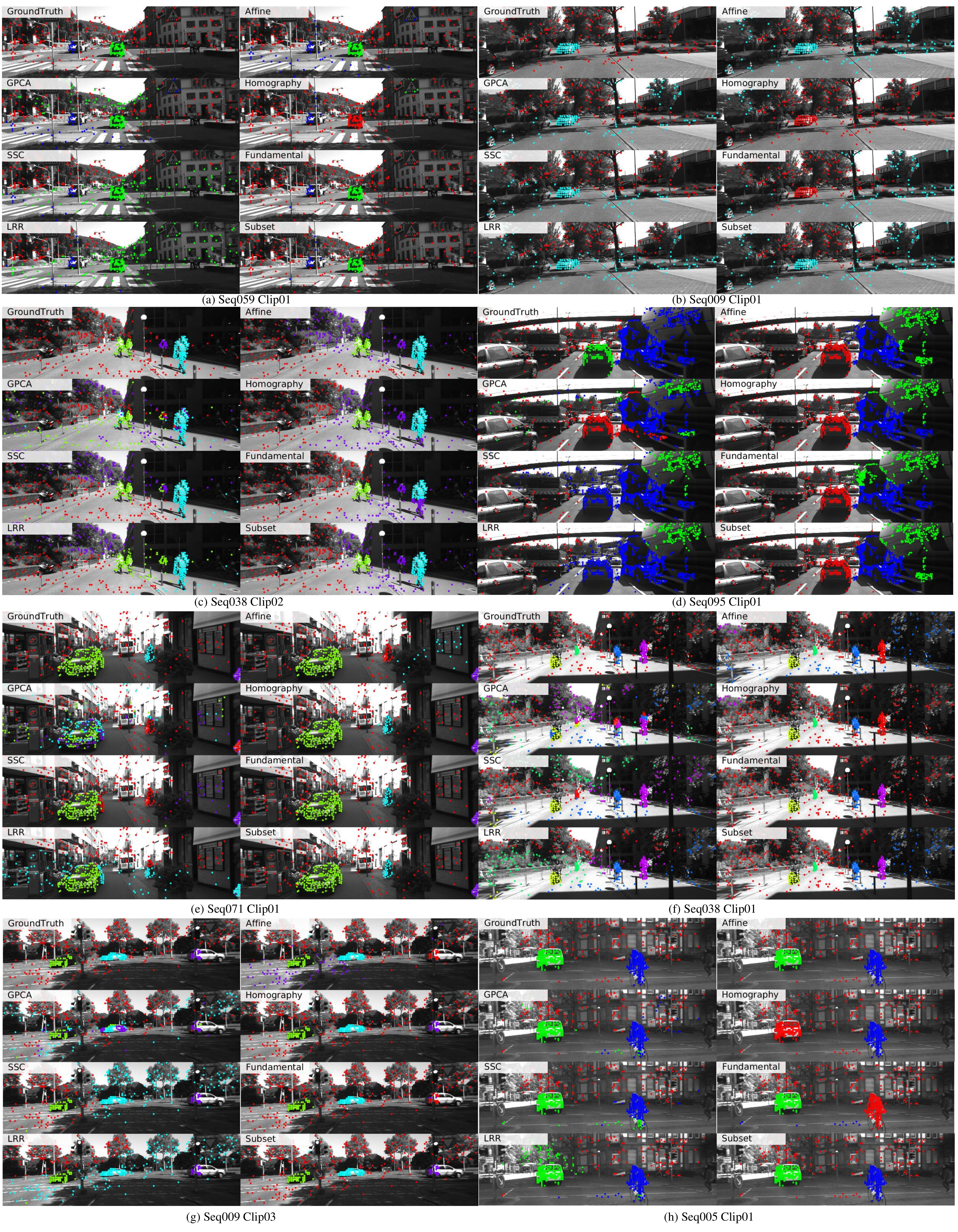}
\caption{Examples of motion segmentation on KT3DMoSeg sequences.}\label{fig:QualitativeKT3DMoSeg}
\end{center}
\end{figure*}

\noindent\textbf{Model Selection} We have discussed in Section~4 how the reconstruction error term and the normalized cut term can be viewed as a precision-complexity trade-off. In particular, the normalized cut criteria can be viewed in information-theoretic terms \cite{raj2010information} as model complexity. In Fig.~\ref{fig:MdlSelQualitative}, we demonstrate qualitatively the interplay of this fundamental precision-complexity trade-off with an example sequence from Hopkins155. The original affinity matrix is presented in Fig.~\ref{fig:MdlSelQualitative} (a) and the reconstructed affinity as explained in Eq~(\ref{eq:RecAff}) from (b) to (h). The red rectangles indicate the ground-truth clusters. The reconstructed affinity matrices in Fig.~\ref{fig:MdlSelQualitative} (b) to (h) are based on clustering the data into 1 to 7 clusters respectively. It is observed from the illustrations that a correct estimate of cluster number results in a good denoising of the raw affinity matrix, as shown in (d). By deviating from the correct model complexity, the resulting models (the reconstructed affinities) suffer from a significant level of deviations from the original affinity matrix. We also plot the normalized cut cost, reconstruction error and the combined final residual against different number of clusters in (i). The normalized cut cost (blue line) goes up monotonically with increasing number of cluster, with increasing speed from 3 clusters onward. This echoes our foregoing analysis in Section~4.2 that the normalized cut cost serves as the complexity penalty term for model selection. We further note that the data fidelity term of reconstruction error (orange line) experiences a steep drop from 1 to 3 clusters and then stays relatively stable when the number of cluster further increases (Fig.~\ref{fig:MdlSelQualitative}(e) to (h)). Given a good combination of the data fidelity and model complexity terms (the final residual as the yellow line), a local minimal is found at number of cluster$=3$, balancing the precision-complexity trade-off.

\begin{figure}
\begin{center}
{\includegraphics[width=1\linewidth]{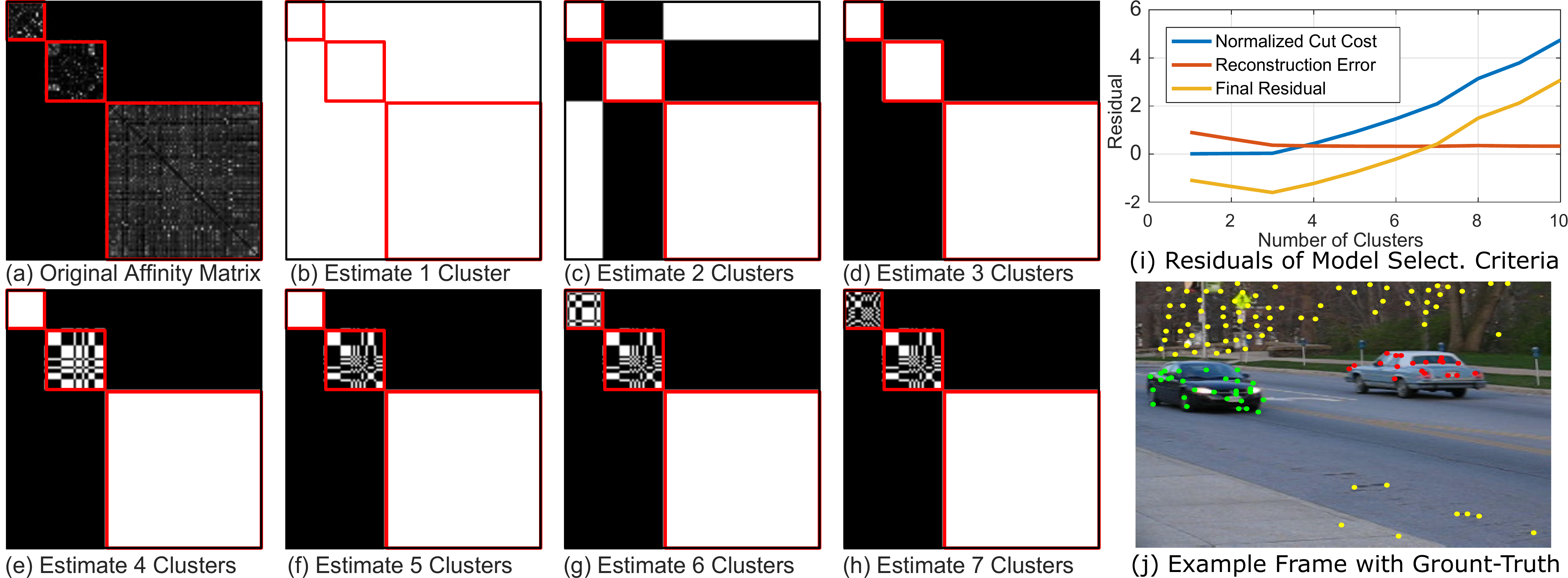}}
\caption{Original affinity matrix v.s. reconstructed affinity and  residuals of model selection criteria.}\label{fig:MdlSelQualitative}
\vspace{-0.5cm}
\end{center}
\end{figure}

\noindent\changed{\textbf{Dense \& Articulated Motion Segmentation} We present motion segmentation examples on the FBMS59 dataset in Fig.~\ref{fig:FBMS59}. We compare with the results obtained by SpectralClustering (Ochs) \cite{ochs2014segmentation} and MultiCut\cite{keuper2015motion} on the first frame of 6 selected sequences. We observe consistently better motion segmentation results by our multi-view fusion approach. For the `dogs01', `farm01' and 'lion01' sequences, our algorithm detects the whole foreground objects while the Ochs\cite{ochs2014segmentation} and MultiCut\cite{keuper2015motion} more or less over-segment the articulated foregrounds. For the `bear02', `meerkat01' and `duck01' sequences, we observe consistently higher recall, i.e. more coverage of the foreground objects, by our algorithm. This observation coincides with the quantitative results (Tab.~\ref{tab:FBMS59}) that our algorithm is on average higher in recall. It is worth noting that for many sequences, all algorithms miss some objects, e.g. the horse in the stable of `farm01' and the wood piece in `lion01', or fail to further separate individual ducks of `duck01' due to insufficient motion difference. In general, our multi-view clustering approach delivers very competitive results on the challenging FBMS59 motion segmentation task, achieving the best overall performance (i.e., F-measure) rather than being top in precision or recall.\\
We further compare the motion segmentation results on the ComplexBackground dataset in Fig.~\ref{fig:ComplexBackground}. For each sequence, we attach the Precision(P), Recall(R) and F-measure(F) for the three approaches, Ochs\cite{ochs2014segmentation}, MultiCut\cite{keuper2015motion} and Ours. We observe good precision for each sequence by our method, while the recalls for `store' and `traffic' are particularly low for all methods. This is due to the large number of small objects missed by all methods. We believe that most of these tiny objects fail to be picked up by these geometric methods as there are not even sufficient feature points lying on the objects. We also notice that both \cite{ochs2014segmentation} and \cite{keuper2015motion} tend to over-segment the scene, e.g. `forest' and `traffic', due to the underlying translational motion model. And occasionally, such over-segmentation could lead to higher precision measure, e.g., in the `forest' scene.
}

\begin{figure*}[htb]
\centering
\includegraphics[width=1.01\linewidth]{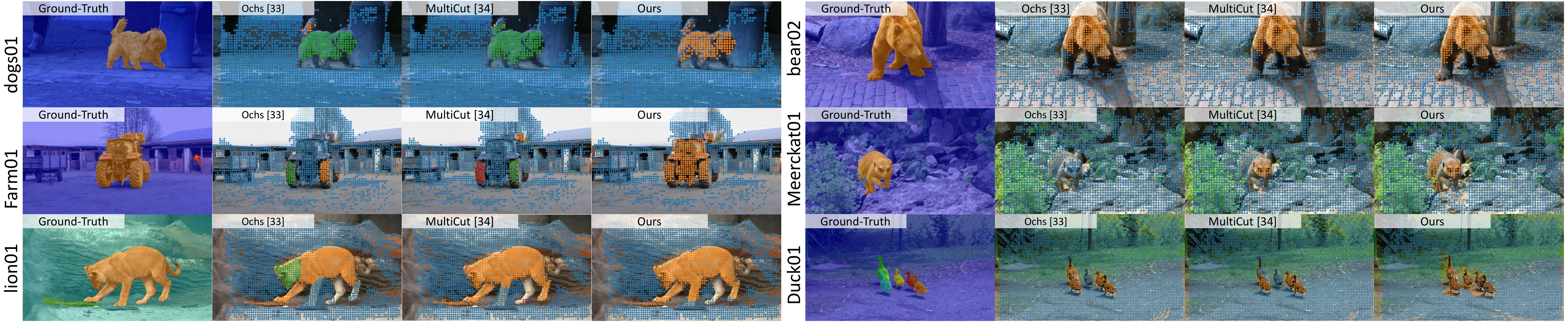}\vspace{-0.2cm}
\caption{Qualitative examples of 6 sequences selected from FBMS59. }\label{fig:FBMS59}
\vspace{-0.6cm}
\end{figure*}

\begin{figure}
\centering
\includegraphics[width=1.0\linewidth]{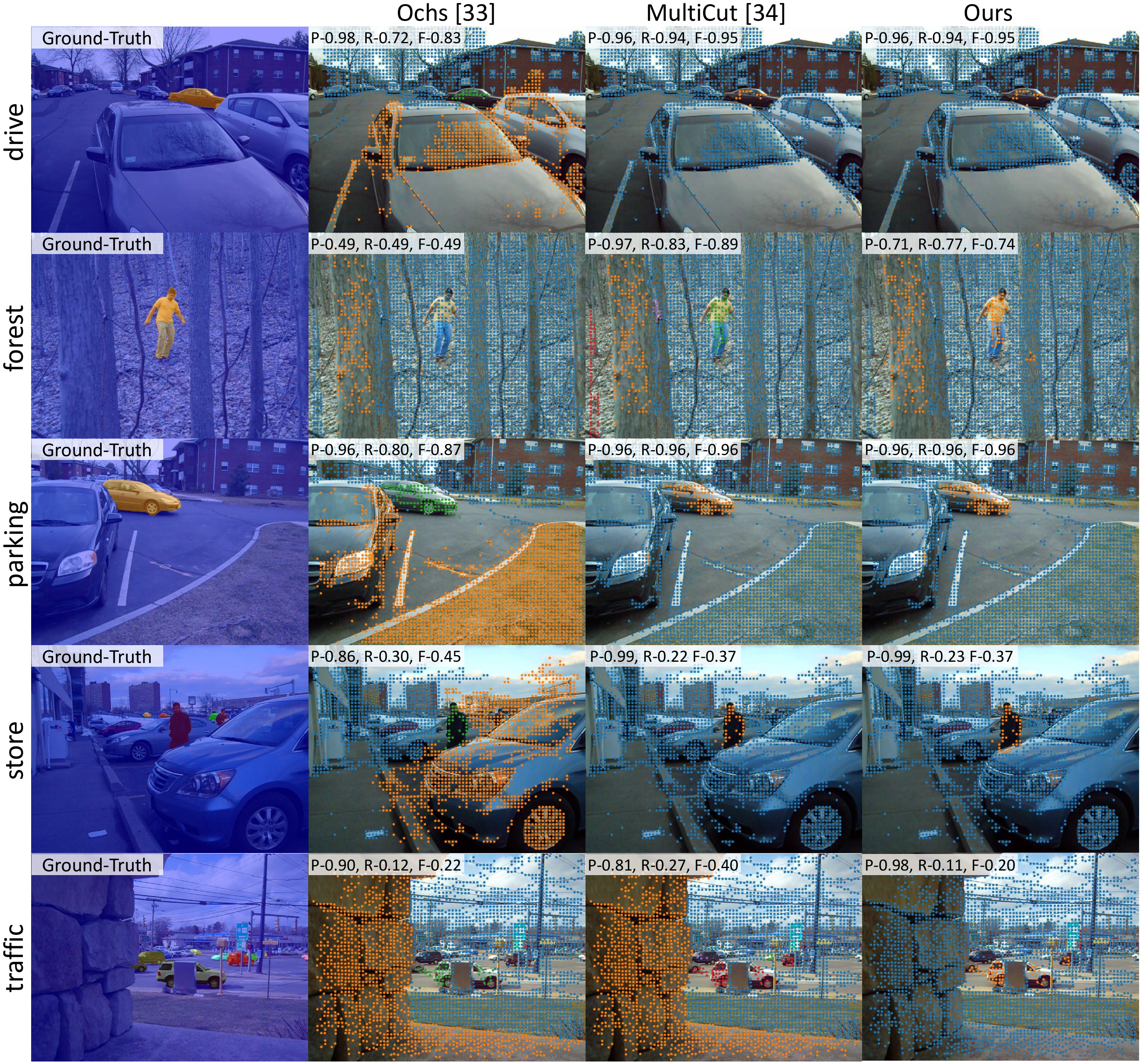}\vspace{-0.2cm}
\caption{Qualitative examples of all 5 sequences of the ComplexBackground dataset.
\vspace{-0.5cm}}\label{fig:ComplexBackground}
\end{figure}

\subsection{Individual Models}\label{sect:IndividualViews}

In this section, we investigate the impact of each individual model by looking into the fusion impact on individual models and the importance of including certain models on the final results.

\noindent\textbf{Fusion Impact on Individual Models} As a result of the co-regularization, each of the geometric models has their views modified; we call these the F-view, H-view, and A-view. We now analyze the performance gain experienced by these models. In particular, we investigate the performance of motion segmentation with the spectral embedding of these models after co-regularization. This is equivalent to using just a single $\matr{U}=\matr{U}_v$ for k-means clustering in the last step of Algorithm~\ref{alg:Subset}. The classification error over all KT3DMoSeg sequences v.s. $\lambda$ and $\gamma$ are presented in Fig.~\ref{fig:KTMoSeg_PerfPerSeq} (b-c). We observe from this evaluation that while the F-view (purple line) does not necessarily produce the best result compared with the H-view, under certain range of $\lambda$ (corresponding to different coerciveness of the co-regularization), the F-view can be corrected so that its full potential is realized, producing the best of all results.

\noindent\textbf{Importance of Individual Models} The importance of each individual model in motion segmentation is determined by the nature of the scene, e.g. perspective effect and camera motion. As we have pointed out, both the affine and homography models have the ability to model weak perspective scenes and rotation-dominant camera motion, whereas the fundamental matrix has the advantage in modeling strong perspective scene. We investigate here if certain models are preferred by each dataset. In particular, we compare the following two-model fusion approaches, affine model with homography ($\matr{A}+\matr{H}$) and homography with fundamental matrix ($\matr{H}+\matr{F}$), for model selection task on three datasets, Hopkins155, MTPV62 and KT3DMoSeg. The results are presented in Table~\ref{tab:ViewImportance}. It is observed from these results that for Hopkins155 and MTPV 62 \footnote{Though a small subset of MTPV62 has stronger perspective effect with camera translation, we observe most of the sequences are captured by hand-held camera with camera rotation being dominant.} where either the perspective effect is weak and/or camera mainly undergoes a rotation-dominant motion, the combination of $\matr{A}+\matr{H}$ outperforms $\matr{H}+\matr{F}$ while the difference between $\matr{A}+\matr{H}$ and combining all three models are smaller. The situation for KT3DMoSeg is reversed: we observe that combining $\matr{H}$ and $\matr{F}$ is optimal, owing to the strong perspective effect and the greater depth complexity in the dataset. In general, choosing the most appropriate models to fuse can make a difference; however, without prior knowledge on the nature of the scene and camera motion, our three-model fusion scheme is still very robust and produces very competitive performances.

{\subsection{Runtime Evaluation}}

{Finally, we evaluate the runtime of our algorithms. Specifically, we separately evaluate the three parts of our model, hypothesis generation, kernel computation and single-model/multi-model clustering. We implement the code in Matlab R2015b and evaluate on a server with two 10 core CPUs (E5-2640) and 128GB memory. Both the mean and standard deviation of the time consumed over all 22 sequences are reported in Tab.~\ref{tab:Runtime}. The overall average times for completing motion segmentation with single motion model are 5.4, 8.0 and 9.2 seconds for Affine, Homography and Fundamental Matrix respectively. The average times for multi-model motion segmentation are 22.2, 24.9 and 23.0 seconds for Kernel Addition (K.A.), Co-Regularization (CoReg) and Subset Constrained Clustering (Subset), respectively. The average number of frames is 17.5 for all 22 sequences at 10 fps. Though the un-optimized code cannot achieve real-time performance, we believe that a more efficient hypothesis generation can significantly reduce the whole processing time. }

\begin{table}[htbp]
  \centering
  \caption{  {
Evaluation of runtime on each component of our pipeline. Time is in seconds.}}
    \resizebox{1\linewidth}{!}{
    {
    \begin{tabular}{p{2em}p{2.5em}p{2.5em}p{2.5em}p{2.5em}p{2.5em}p{2.5em}p{2.5em}p{2.5em}}
    \toprule
    Steps & \multicolumn{3}{c}{Hypothesis} & \multicolumn{1}{c}{Kernel} & \multicolumn{4}{c}{Clustering} \\
    \cmidrule(lr){2-4} \cmidrule(lr){5-5} \cmidrule(lr){6-9}
    Model & \multicolumn{1}{c}{Affine} & \multicolumn{1}{c}{Homo} & \multicolumn{1}{c}{Fund} & \multicolumn{1}{c}{A/H/F} & \multicolumn{1}{c}{A/H/F} & \multicolumn{1}{c}{K.A.}& \multicolumn{1}{c}{CoReg} & \multicolumn{1}{c}{Subset} \\
    \midrule
    Time  & 4.3$\pm$1.2 & 6.9$\pm$1.9 & 8.1$\pm$4.2 &  0.9$\pm$1.0  &   0.2$\pm$0.1  & 0.2$\pm$0.1 &   2.9$\pm$1.0  & 1.0$\pm$1.4 \\
    \bottomrule
    \end{tabular}%
    }
    }
  \label{tab:Runtime}%
  \vspace{-0.5cm}
\end{table}%

\begin{table}[htbp]
  \centering
\caption{The impact of different combinations of multi-model fusion schemes on model selection task. The correct rate ($\%$) and mean error ($\%$) are separated by a slash.}
\resizebox{0.82\linewidth}{!}{
    \begin{tabular}{lllll}
    \toprule
          &       & \multicolumn{3}{c}{Correct Rate / Mean Error (\%)} \\
\cmidrule{3-5}    Dataset & Models & KerAdd & CoReg & Subset \\
    \midrule
    \multirow{3}[2]{*}{Hopkins155} & \textbf{A+H} & 90.32 / 2.24 & 90.32 / 2.19 & 90.32 / 2.31 \\
          & \textbf{H+F} & 80.65 / 4.54 & 80.00 / 4.40 & 80.00 / 4.46 \\
          & \textbf{A+H+F} & 89.03 / 3.44 & 91.61 / 1.95 & 91.61 / 1.83 \\
    \midrule
    \multirow{3}[2]{*}{MTPV62} & \textbf{A+H} & 83.87/ 3.80 & 87.10/ 2.91 & 87.10/ 4.75 \\
          & \textbf{H+F} & 85.48 / 3.90 & 72.58 / 6.63 & 72.58 / 6.51 \\
          & \textbf{A+H+F} & 85.48 / 4.21 & 83.87 / 4.87 & 83.87 / 3.23 \\
    \midrule
    \multirow{3}[2]{*}{KT3DMoSeg} & \textbf{A+H} & 40.91 / 35.32 & 40.91 / 31.74 & 31.82 / 34.18 \\
          & \textbf{H+F} & 63.64 / 9.61 & 63.64 / 6.91 & 68.18 / 8.67 \\
          & \textbf{A+H+F} & 68.18 / 7.51 & 63.64 / 11.33 & 68.18 / 14.74 \\
    \bottomrule
    \end{tabular}%
    }
  \label{tab:ViewImportance}%
\end{table}%

\section{Conclusion}
In this paper, we have contributed to an understanding of the strengths and drawbacks of homography and fundamental matrices as a geometric model for motion segmentation, not only in the extant datasets such as Hopkins155, but also for real-world sequences in KT3DMoSeg. Not only do we account for the unexpected success of the homography approach when the affinities are accumulated to over all slicing planes, we also reveal its real limitation in real-world scenes. The geometrical exactness of the fundamental matrix approach is theoretically appealing; we show how its potential can be harnessed in a multi-model spectral clustering fusion scheme. Given kernels induced from multiple types of geometric models, we evaluate several techniques to synergistically fuse them. 
For model selection, we propose to tradeoff the data fidelity and model complexity in our NCRE scheme. \changed{Finally, we carry out experiments on Hopkins155, Hopkins12, MTPV62, FBMS59 and ComplexBackground, on most of which achieving state-of-the-art performances for both motion segmentation and model selection.} In light of the demand for real-world motion segmentation, we further propose a new dataset, the KT3DMoSeg dataset, to reflect and investigate real challenges in motion segmentation in the wild.

\section{Acknowledgement}
This work was supported by the Singapore PSF grant 1521200082.



%

%

\begin{IEEEbiography}[{\includegraphics[width=1in,height=1.25in,clip,keepaspectratio]{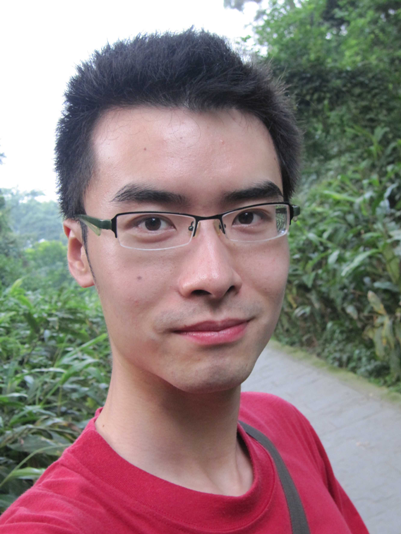}}]{Xun Xu}
received the B.E. degree from Sichuan
University, in 2010 and the PhD degree from Queen Mary University of London in 2016. He is currently
a research fellow with the Electrical and Computer Engineering Department of National University of Singapore.
His research interests include motion segmentation, surveillance video
understanding, zero-shot learning, and action recognition.
\end{IEEEbiography}

\begin{IEEEbiography}
[{\includegraphics[width=1in,height=1.25in,clip,keepaspectratio]{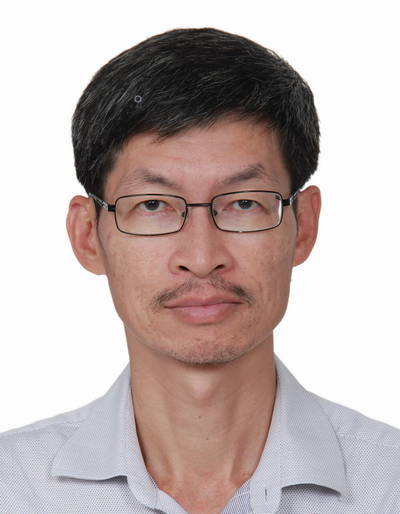}}]{Loong-Fah Cheong}
Loong-Fah Cheong received the BEng degree from the National University of Singapore, and the PhD degree from the University of Mary- land at College Park, Center for Automation Research, in 1990 and 1996, respectively. In 1996, he joined the Department of Electrical and Computer Engineering, National University of Singapore, where he is an associate professor currently. His research interests include the processes in the perception of three-dimensional motion, shape, and their relationship, as well as the 3D
motion segmentation and the change detection problems.
\end{IEEEbiography}


\begin{IEEEbiography}[{\includegraphics[width=1in,height=1.25in,clip,keepaspectratio]{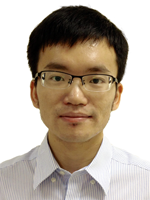}}]{Zhuwen Li}
received the B.E. in Computer Science from Tianjin University in 2008, the Master's degree in Computer Science from Zhejiang University in 2011, and the Ph.D. degree in Electrical and Computer Engineering from National University of Singapore in 2014. Currently, he is a Postdoc Researcher at Intel Intelligent Systems Lab. His research interests include motion analysis, 3D vision and graph neural networks.
\end{IEEEbiography}

%




\end{document}